\PassOptionsToPackage{table}{xcolor}
\RequirePackage{amsmath}
\RequirePackage{amsfonts}

\documentclass{fairmeta}
\usepackage{newpxtext,newpxmath}
\renewcommand{\title}[1]{\newcommand{\titlelist}{{\fontsize{24}{28}\selectfont\sffamily\bfseries\color{clmblue} #1}}}

\usepackage{amsmath,amsfonts,bm}

\def\eqref#1{equation~\ref{#1}}

\def\1{\bm{1}}

\DeclareMathAlphabet{\mathsfit}{\encodingdefault}{\sfdefault}{m}{sl}
\SetMathAlphabet{\mathsfit}{bold}{\encodingdefault}{\sfdefault}{bx}{n}

\usepackage{xspace}
\usepackage{verbatim}
\usepackage{booktabs} 
\usepackage{arydshln} 
\usepackage{wrapfig} 
\usepackage{mathtools}

\usepackage{listings}
\lstdefinestyle{drctx}{
  language=Python,
  basicstyle=\ttfamily\small,
  keywordstyle=\ttfamily,
  commentstyle=\color{gray},
  showstringspaces=false,
  keepspaces=true,
  columns=fixed,
  basewidth=0.53em,
  escapeinside={(*@}{@*)},
  xleftmargin=0pt,
  aboveskip=2pt,
  belowskip=0pt,
  frame=none,
}
\definecolor{ctxhl}{HTML}{C92A2A}
\definecolor{evoblue}{HTML}{1F77B4}
\definecolor{evopurple}{HTML}{7B3FA0}

\definecolor{readhl}{HTML}{2B8A3E}

\definecolor{clmblue}{HTML}{0668E1}
\newcommand{\clmhl}[1]{\textcolor{clmblue}{\textbf{#1}}}
\newcommand{\skl}{{\color{clmblue}s}}
\definecolor{ctxorigfill}{HTML}{BCD8EE}\definecolor{ctxorigedge}{HTML}{0072B2}\definecolor{ctxorigtext}{HTML}{0B3D6B}
\definecolor{ctxeditfill}{HTML}{F4C8D6}\definecolor{ctxeditedge}{HTML}{C23A63}\definecolor{ctxedittext}{HTML}{7A2040}
\newcommand{\ctxboxgeneric}[4]{\tikz[baseline=(ctxn.base)]\node[draw=#1,fill=#2,text=#3,rounded corners=1.6pt,line width=0.45pt,inner xsep=2.6pt,inner ysep=1.4pt,outer sep=0pt](ctxn){\ensuremath{#4}};}
\newcommand{\ctxorig}[1]{\ctxboxgeneric{ctxorigedge}{ctxorigfill}{ctxorigtext}{#1}}
\newcommand{\ctxedit}[1]{\ctxboxgeneric{ctxeditedge}{ctxeditfill}{ctxedittext}{#1}}

\colorlet{metablue}{clmblue}

\usepackage{tocloft}
\addtocontents{toc}{\protect\setcounter{tocdepth}{-1}}
\usepackage{csquotes}
\newtcolorbox{quotebox}{
  enhanced,
  breakable,
  colback=gray!5,
  colframe=gray!35,
  boxrule=0.5pt,
  arc=2pt,
  left=8pt,
  right=8pt,
  top=6pt,
  bottom=6pt
}

\usepackage{wrapfig}
\usepackage{threeparttable} \usepackage{array}

\usepackage{color-edits}
\definecolor{oc-orange-9}{HTML}{D9480F}
\definecolor{oc-cyan-8}{HTML}{0C8599}
\definecolor{rs-color}{HTML}{e87ea1}
\definecolor{forestgreen}{HTML}{009B55}
\addauthor{ss}{oc-orange-9}
\addauthor{rs}{rs-color}

\newtcolorbox{msgbox}[2][gray]{
    enhanced, breakable,
    colback=#1!6, colframe=#1!50!black!40,
    boxrule=0.5pt, arc=2pt, left=6pt, right=6pt, top=3pt, bottom=3pt,
    title={\scriptsize\sffamily\bfseries #2}, coltitle=black,
    colbacktitle=#1!15, toptitle=1.5pt, bottomtitle=1.5pt,
    before skip=4pt, after skip=4pt
}

\colorlet{swehl}{metabg}
\colorlet{storagepink}{metabg}

\usepackage{amsmath}

\newcommand{\ours}{CLM\xspace}
\newcommand{\oursplural}{CLMs\xspace}

\newcommand{\OURSplural}{Context Language Models\xspace}

\newcommand{\ourbench}{ContextBench\xspace}

\definecolor{tengbrown}{HTML}{7A4E2D}

\usepackage{tikz}

\title{\OURSplural}

\author[1,2]{Rulin~Shao}
\author[3]{Shannon~Zejiang~Shen}
\author[1,2]{Junjie~Oscar~Yin}
\author[1]{Yuetai~Li}
\author[1]{Minheng~Wang}
\author[1]{Hamish~Ivison}
\author[1]{Radha~Poovendran}
\author[4]{Nathan~Lambert}
\author[1]{Teng~Xiao}
\author[2]{Mike~Lewis}
\author[2]{Wen-tau~Yih}
\author[1,2]{Luke~Zettlemoyer}
\author[1]{Pang~Wei~Koh}

\affiliation[1]{University of Washington}
\affiliation[2]{Meta Superintelligence Labs}
\affiliation[3]{MIT}
\affiliation[4]{Trillium Labs}

\date{\today}
\correspondence{Rulin Shao at \email{rulin@cs.washington.edu}}
\metadata[Code]{\url{https://github.com/facebookresearch/context-language-models}}

\abstract{
We introduce \clmhl{\OURSplural (\oursplural)}, language models that natively manage their own context. We implement this by treating the \clmhl{context as a file} and allowing the model to make unrestricted updates to this file. This allows the model to learn what is most important to maintain in context, and naturally extends to multi-agent systems where multiple agent contexts coexist as files. Building \oursplural \clmhl{zero-shot} with existing models outperforms SOTA context-management strategies across a variety of tasks: 11.4\% higher accuracy with 21.5\% fewer FLOPs on BrowseComp-Plus, 5\% higher scores with 59\% fewer FLOPs on 12-hour EdgeBench, and 65\% greater improvement with the same compute on a 24-hour multi-repository agent-swarm task.
Moreover, by shifting context management from external harness control to intrinsic model behavior, \oursplural naturally enable both \clmhl{in-context and parametric learning} of context-management strategies.
We show that \oursplural can be steered with natural-language instructions evolved through a standard skill-optimization loop, improving held-out accuracy by up to 35.9 points on a context-management task while reducing compute. 
We also introduce an online reinforcement learning method for \oursplural, improving Qwen3.5-9B performance on BrowseComp-Plus by 47.6\% while using 12\% fewer FLOPs.
Finally, we co-design \clmhl{Suffix Cache Reuse} for \ours{}
serving, further reducing server-side compute by 35\% relative to standard SGLang at matched performance.
}

\usepackage{placeins}
\begin{document}

\maketitle

\section{Introduction}\label{sec:intro}

Despite the fact that context is the cornerstone that allows a language model (LM) to process and retain information over time,  
context management is not traditionally a native LM capability.
Instead, prior work mostly relies on harnesses, either hand-engineered~\citep{cassano2026trainingcomposer,openai2026codex,merrill2026terminal} or optimized offline by agents~\citep{lee2026meta}.
Recent work adds a constrained set of tools with fixed strategies such as compaction, offloading, and retrieval, to the agent's action space~\citep{yan2026memory,yu2026agentic,li2026acm,liu2026context,zhang2025recursive}. 
In contrast, we show that giving LMs unrestricted access to manage their own context outperforms human-designed baselines, enabling adaptive and creative context-management strategies to emerge.
Our findings echo The Bitter Lesson~\citep{sutton2019bitter}: we should let LMs search for and learn better strategies that go far beyond existing human priors.

\begin{figure}[t]
    \centering
    \includegraphics[width=\linewidth]{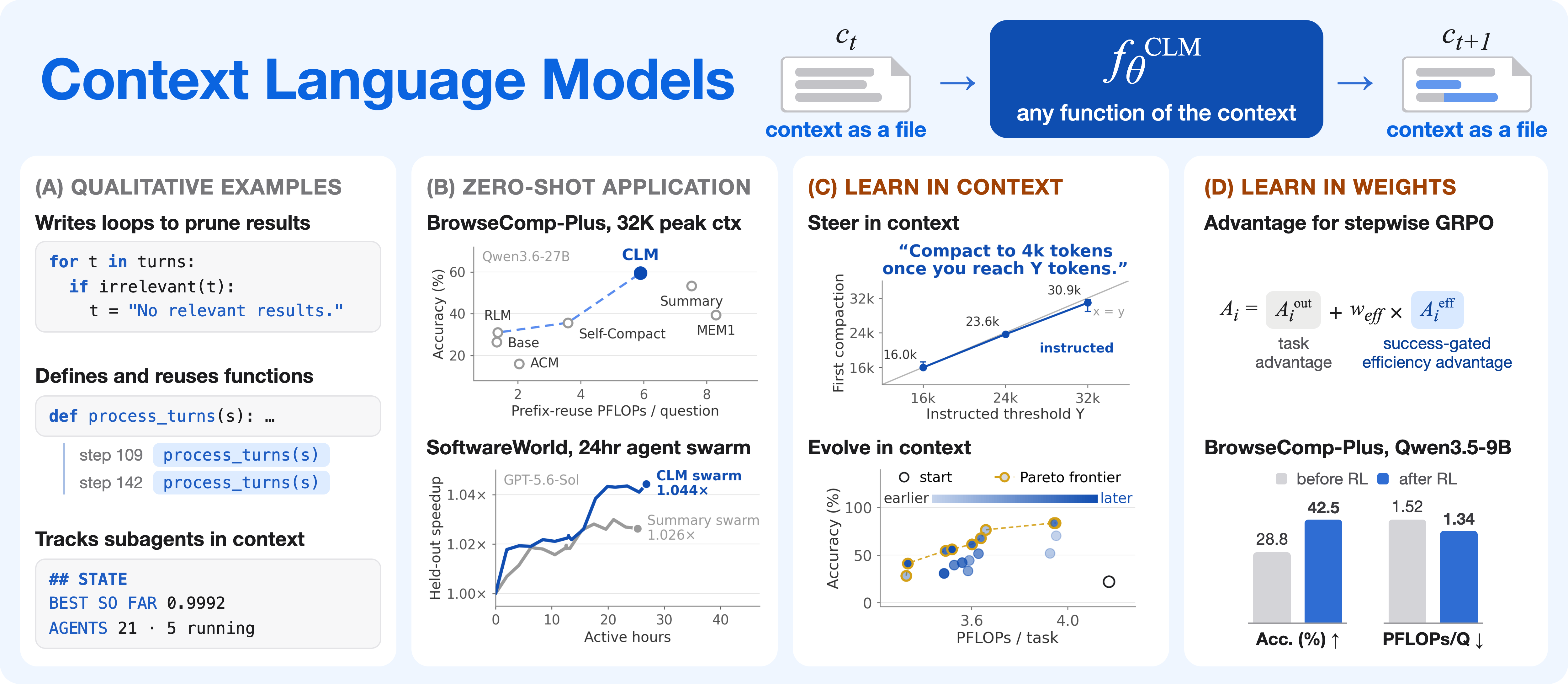}
    \caption{
\textbf{\OURSplural (\oursplural) natively manage their own context by treating context as a file.}
\oursplural{} work out of the box and can be further improved through in-context learning and reinforcement learning. 
\textcolor{clmblue}{(a)} Qualitative examples of creative context-management behaviors introduced by \ours{}.
\textcolor{clmblue}{(b)} Out of the box, \oursplural{} improve performance at lower cost on BrowseComp-Plus, a deep-research benchmark, and Software World, where an agent swarm jointly optimizes six interdependent repositories.
\textcolor{clmblue}{(c)} \oursplural{} can follow textual instructions to adopt corresponding context-management strategies (top) or evolve better strategies through a skill-evolution loop on \ourbench{} (bottom).
\textcolor{clmblue}{(d)} \oursplural{} can explore and internalize context-management strategies through online reinforcement learning. By using a success-gated efficiency advantage for stepwise GRPO, we improve both accuracy and efficiency for \oursplural{} simultaneously.
}
    \label{fig:teaser}
\end{figure}

Concretely, we introduce \clmhl{\OURSplural{} (\oursplural{})}, which are natively capable of managing their own context.
We show existing LMs can be turned into strong \oursplural 
and can be further improved through in-context learning and reinforcement learning.
Formally, a \ours parametrized by \(\theta\) makes context an artifact of the LM: 
$c_{t+1} = f^{\mathrm{CLM}}_\theta(c_t)$, where $c_t$ is the context at turn $t$ and $f^{\mathrm{CLM}}_\theta$ can be an arbitrary function controlled by \oursplural.
In contrast, a standard LM simply appends new tokens to the existing context: 
$c_{t+1} = c_t \oplus f^{\mathrm{LM}}_\theta(c_t)$.

We implement \oursplural by treating \clmhl{context as a file}.
Specifically, we mirror the context into a storage space with LM write access. The LM can either append newly generated tokens or use Bash  to freely edit the context file, with each modification immediately synchronized to the LM's live context for the next turn.
This design naturally extends to multi-agent systems, where multiple context files can coexist and be managed by \oursplural for \clmhl{agent-swarm} or \clmhl{subagent} workloads.

Arbitrary context edits in \oursplural pose new challenges for existing serving systems, which typically only reuse cached states for matching prefixes, forcing re-prefilling after in-the-middle edits. 
We account for this by introducing \clmhl{prefix-reuse FLOPs}, which capture the trajectory-wide inference costs of decoding, prefilling, and re-prefilling in standard LM serving,
and show that \oursplural{} remain more compute-efficient under standard serving through better context management. 
We further develop \clmhl{Suffix Cache Reuse (SCR)}, which reuses cached states beyond the matching prefix to reduce re-prefilling while empirically preserving task performance.
We also introduce \clmhl{\ourbench} as a diagnostic benchmark that decouples context management from reasoning and knowledge, revealing the limitations of existing context management methods.

We show that \oursplural, applied \clmhl{zero-shot} to models like Qwen3.6-27B and GPT5.6-Sol, outperform existing baselines and task-specific harnesses across diverse long-horizon tasks, ranging from hundreds to thousands of turns and up to 24 hours of runtime.
Compared with existing harness-defined and action-based methods, \ours{} achieves 11.4\% higher accuracy with 21.5\% fewer prefix-reuse FLOPs than the strongest baseline on the deep-research benchmark BrowseComp-Plus~\citep{chen2025browsecomp}, while matching the strongest baseline's accuracy on the terminal-coding benchmark TerminalBench~2.1~\citep{merrill2026terminal} with 29.5\% fewer FLOPs.
On mathematical optimization tasks, \ours{} outperforms specialized evolutionary harnesses such as OpenEvolve~\citep{openevolve} by up to 16.8\% (Heilbronn) and 3.0\% (circle packing).
On long-running software optimization, \ours{} outperforms 
Codex-style summarization: on 12-hour EdgeBench~\citep{zhu2026edgebench} (a 10-task subset), \ours{} scores 5\% higher while using 59\% fewer prefix-reuse FLOPs, and on a 24-hour six-repository agent-swarm task, it achieves 65\% greater end-to-end speedup at the same compute.
Moreover, when Suffix Cache Reuse is further applied, it helps reduce server-side compute by 35\% with matched performance compared with standard SGLang serving.
Qualitatively, we find that \oursplural{} come up with novel emergent behaviors such as defining and maintaining trackers for multi-agent orchestration, introducing a new chat role for internal notes, and defining reusable context-management functions.

By shifting context management from external harness control to intrinsic model behavior, \oursplural{} naturally enable \clmhl{in-context learning} and \clmhl{parametric learning} for context management.
We first show that users can \clmhl{steer} context management simply by telling the agent their desired strategy. 
In addition, \oursplural can \clmhl{evolve} an in-context skill document that captures useful context-management procedures for future reuse, 
improving held-out accuracy on \ourbench{} by up to 35.9 points at lower compute.
For training, we introduce a \clmhl{success-gated efficiency advantage} in stepwise GRPO~\citep{shao2024deepseekmath} that rewards efficient \ours{} trajectories among successful ones.
Training Qwen3.5-9B on deep research tasks in this manner  improves \ours{} from 28.8\% to 42.5\% on BrowseComp-Plus, outperforming a Codex-style summary harness trained with the same recipe by 0.4 points while using 38.8\% fewer FLOPs.
Overall, we show that by treating context management as a native LM capability, \oursplural enable more effective and efficient strategies to be searched for and learned.

\section{Related Work}
\label{sec:related_work}

\textbf{From harness-defined to action-based context management.}
Most existing harnesses
compact accumulated histories according to a fixed harness policy~\citep{cassano2026trainingcomposer,openai2026codex,merrill2026terminal,zhou2026mem1}, such as at a predefined length threshold or at every turn.
Recent work gives the model increasing control through human-defined actions:
AutoCompact~\citep{zhang2026autocompact} and Self-Compact~\citep{li2026self} let the model decide when to compact; Context-as-a-Tool~\citep{liu2026context} exposes model-triggered compaction over a predefined portion of the context;
ACM~\citep{li2026acm} adds model-triggered offloading and retrieval; and Sculptor~\citep{li2026sculptor} lets the model select context fragments to operate on.
Across this progression, model autonomy increases but remains restricted to a human-defined action space.
Our work pushes this autonomy to its limit by granting the model full agency over its context.

\textbf{Context as a REPL variable.}
Recursive Language Models (RLMs)~\citep{zhang2025recursive} 
treat a long input as a read-eval-print loop (REPL) variable that LMs can recursively access on demand. 
This addresses when and what information to \textit{read} into the context. 
However, RLMs do not address how the live context itself should be managed. Retrieved information is still appended to the live context, which continues to grow over time.
In contrast, our work makes the live context editable, giving \oursplural full control over their context.

We provide extended related work in Appendix~\ref{app:extended_related_work}, with more detailed comparisons to existing context-management baselines and a discussion of meta-harness optimization, reinforcement learning, cache reuse, and the relationship between context management and external memory.

\section{Pilot Study with \ourbench: A Diagnostic for Context Management}

\begin{figure*}[h]
\centering
\hspace*{0.052\linewidth}%
\begin{minipage}[t]{0.2105\linewidth}\centering
{\footnotesize\textbf{Needle Retention}}\\[-1pt]{\scriptsize (selective verbatim retention)}\\[2pt]
\includegraphics[width=\linewidth]{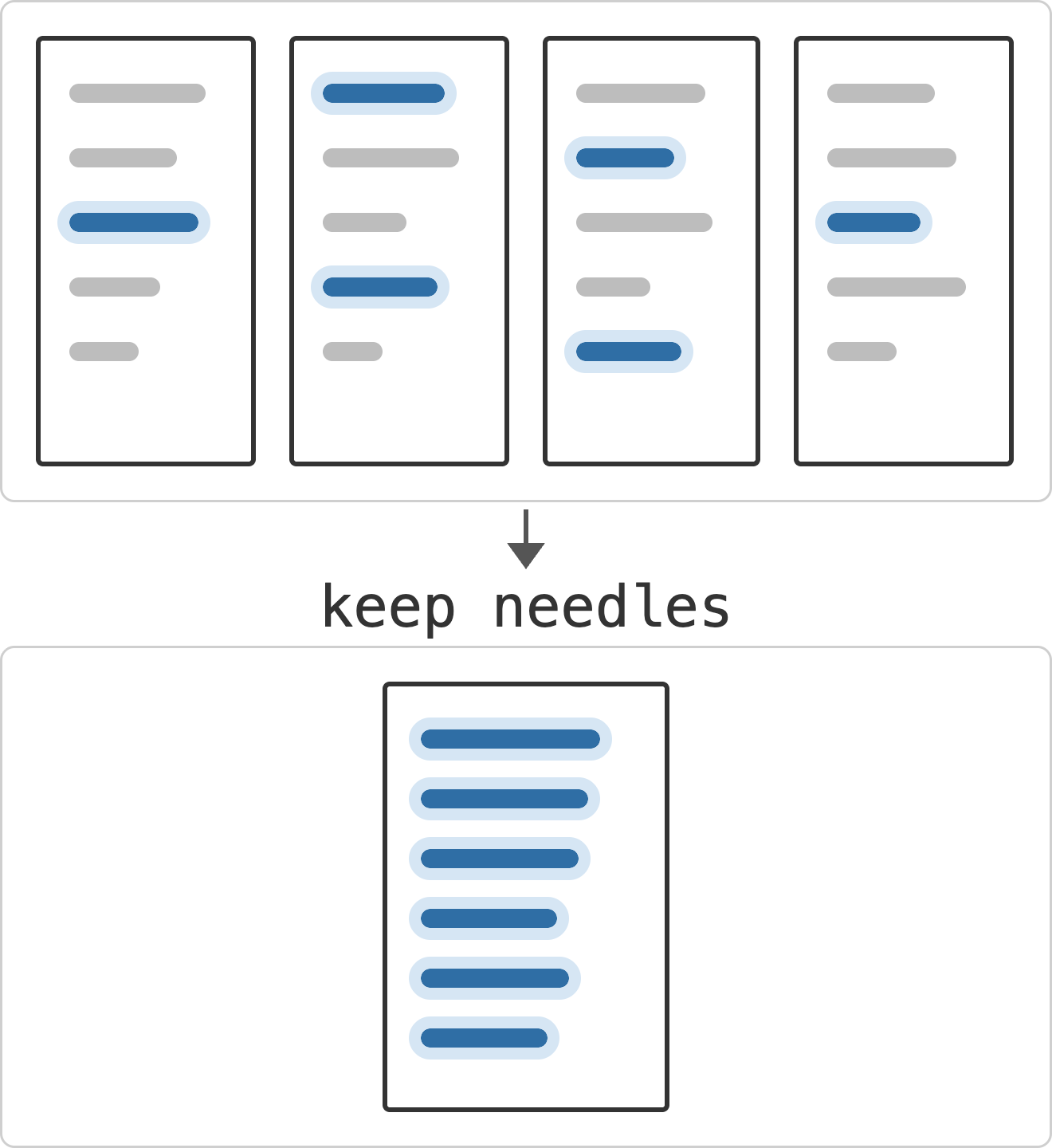}
\end{minipage}%
\hspace{0.0337\linewidth}%
\begin{minipage}[t]{0.2105\linewidth}\centering
{\footnotesize\textbf{Sudoku Sketchpad}}\\[-1pt]{\scriptsize (in-place surgical editing)}\\[2pt]
\includegraphics[width=\linewidth]{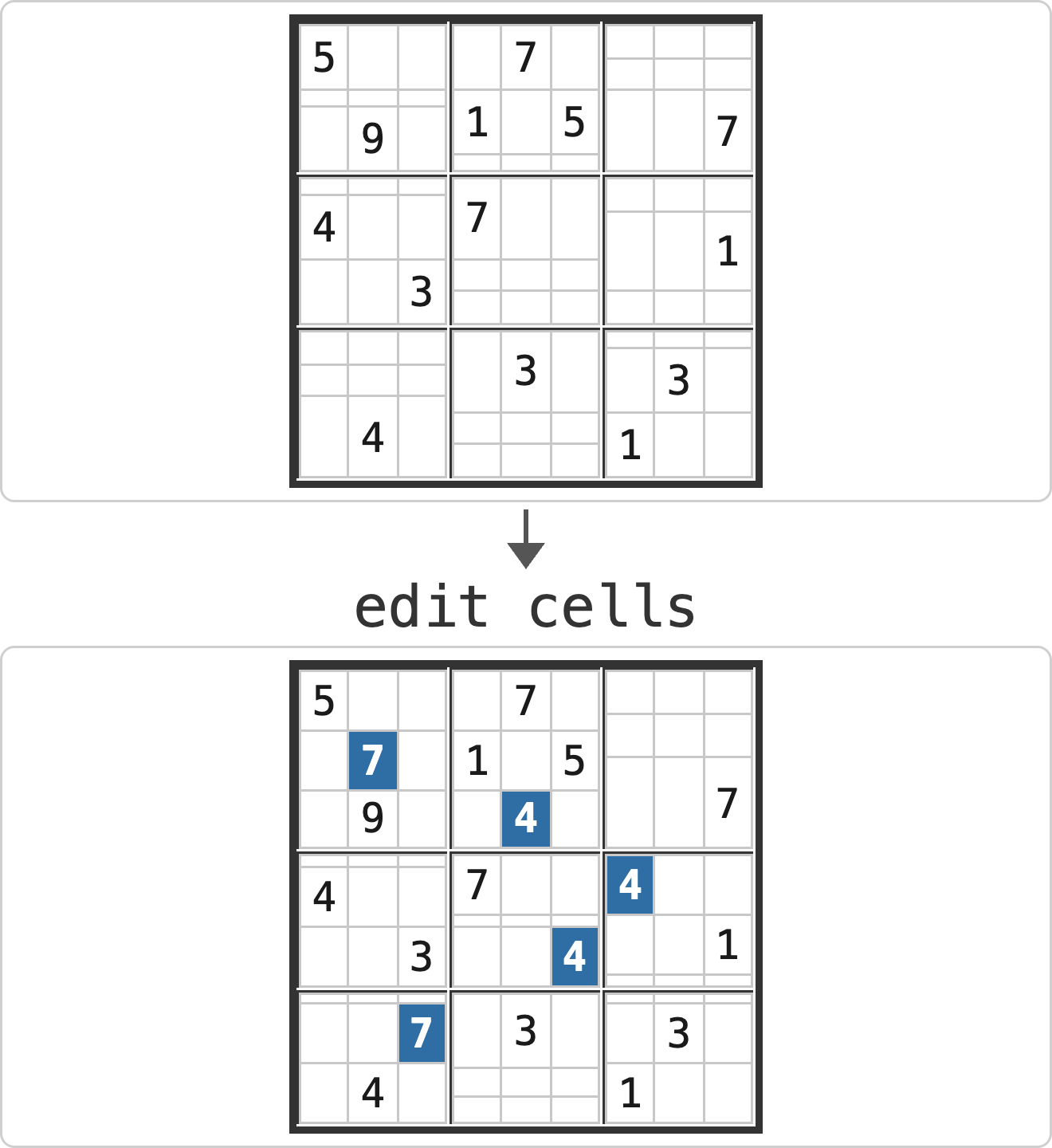}
\end{minipage}%
\hspace{0.0337\linewidth}%
\begin{minipage}[t]{0.2105\linewidth}\centering
{\footnotesize\textbf{KV Store}}\\[-1pt]{\scriptsize (offloading \& retrieval)}\\[2pt]
\includegraphics[width=\linewidth]{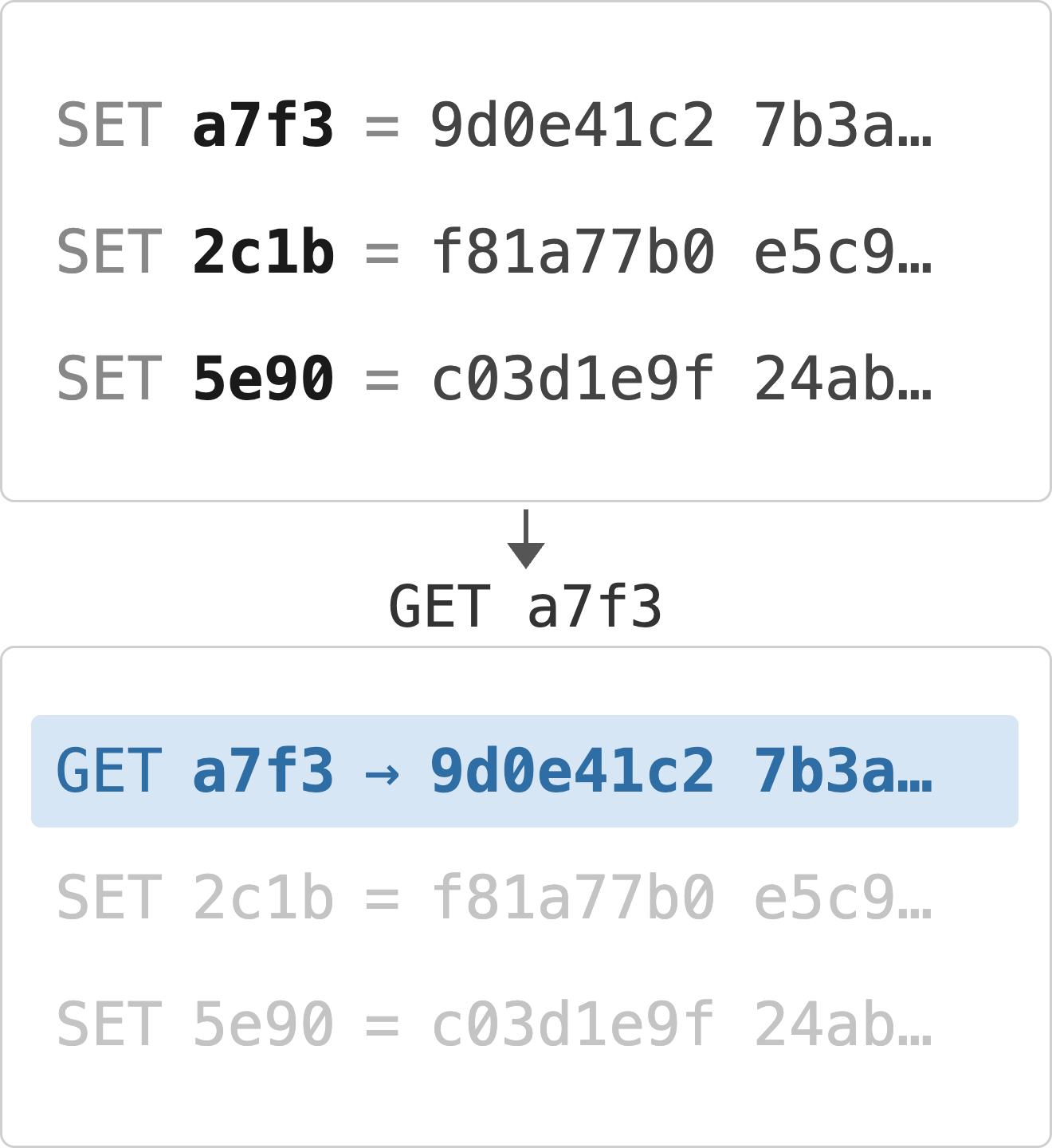}
\end{minipage}%
\hspace{0.0337\linewidth}%
\begin{minipage}[t]{0.2105\linewidth}\centering
{\footnotesize\textbf{Log Triage}}\\[-1pt]{\scriptsize (offloading \& retrieval)}\\[2pt]
\includegraphics[width=\linewidth]{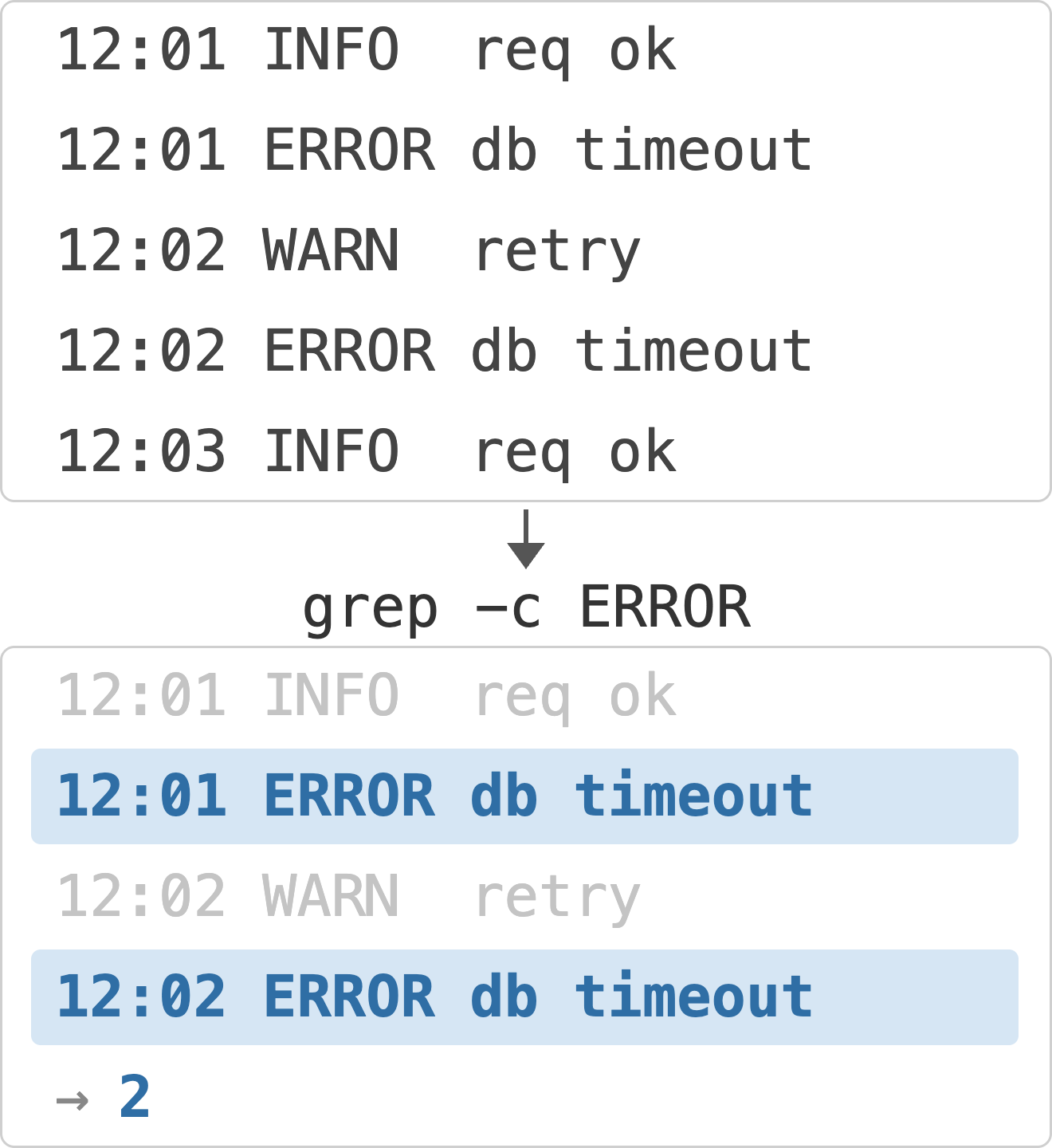}
\end{minipage}%
\hspace*{0.005\linewidth}%
\\[3pt]
\includegraphics[width=\textwidth]{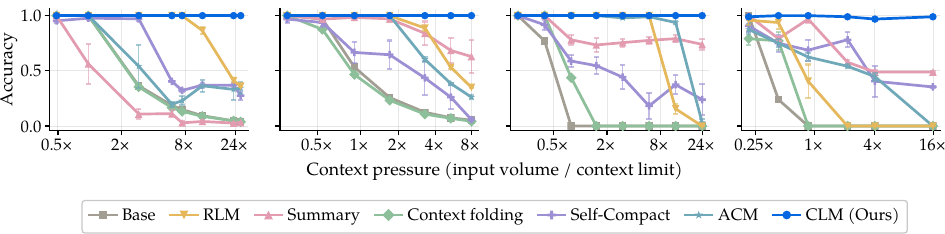}
\caption{
Illustration of the four tasks in \ourbench{} and a performance comparison of \ours{} against baselines using GPT-5.4 with a 32K context limit.
}
\label{fig:synthetic_task_results_main}
\end{figure*}

We start with a pilot study showing how existing context management strategies can fail in simple tasks.
To isolate context management from other reasoning or knowledge capabilities, we developed \clmhl{\ourbench}, a diagnostic evaluation suite with the four synthetic tasks shown in
Figure~\ref{fig:synthetic_task_results_main}: \textbf{Needle Retention} tests selective verbatim retention, simulating the need to preserve important information over time; \textbf{Sudoku Sketchpad} tests surgical in-place updates to the live context by maintaining a Sudoku board as users stream in moves; \textbf{KV Store} and \textbf{Log Triage} test exact recall through offloading and retrieval of massive values and working logs. 
We evaluate \ourbench{} with several context-management strategies, including Mini-SWE-Agent~\citep{yang2024sweagent} (the base harness without context management), Codex-style Summary~\citep{openai2026codex}, Context Folding~\citep{sun2025scaling}, and RLM, Self-Compact, and ACM, as introduced in Section~\ref{sec:related_work}. We also evaluate \ours{}, which will be introduced in Section~\ref{sec:method}.
Details of the evaluation and qualitative examples for \ourbench{} are provided in Appendix~\ref{app:context_bench}.

We fix the context limit at 32K and vary the context pressure (the ratio of input volume to context limit) up to \(24\times\).
The results in Figure~\ref{fig:synthetic_task_results_main} show that these fixed strategies cannot adapt well to the live context: Summary-based compaction can lose or hallucinate information on Needle Retention and Sudoku Sketchpad; methods without flexible in-place editing must regenerate the full Sudoku state for every fine-grained user edit; and standard coding tools can offload information on KV Store and Log Triage but cannot evict it from the live context on demand. As a result, none of the existing methods performs perfectly even on these simple tasks. These failures motivate fully adaptive, model-controlled context management.

\section{\OURSplural (\oursplural)}\label{sec:method}

\subsection{Formal Definition and Implementation with Context as a File}

\clmhl{\OURSplural (\oursplural)} generalize the append-only context transition of a standard LM to a model-controlled context transition.
Standard LMs append model output to the current context:
\begin{equation}
c_{t+1} = c_t \oplus f^{\mathrm{LM}}_\theta(c_t),
\end{equation}
where \(\oplus\) denotes concatenation. 
In contrast, \oursplural delegate full responsibility for maintaining the context to the \ours itself, directly creating the next context:
\begin{equation}\label{eq:clm}
c_{t+1} = f^{\mathrm{CLM}}_\theta(c_t),
\end{equation}
where $f^{\mathrm{CLM}}_\theta$ can be an arbitrary function controlled by \oursplural.
Eq.~\ref{eq:clm} subsumes prior approaches that expose a set of context-management tools through the harness.
However, prior work requires context-management functions to be predefined in the harness. Our work instead makes \oursplural responsible for defining these functions themselves as a \textit{meta-capability}, either implicitly through their planning or explicitly as reusable functions, with one explicit example shown in Figure~\ref{fig:clm-examples}\subref{fig:clm-examples-d}.

\textbf{\clmhl{Context-as-a-file} implementation for \oursplural.}
To implement CLMs, we mirror the LM's live context as a directly editable file and provide its path in the system prompt. 
The LM can edit this file using general Bash commands, just as it would edit other files in storage.
Unlike ordinary files, edits to the context file are automatically synchronized with the LM's context and sent to the LLM server for continued generation.
When the LM does not edit the context file, the generated tokens are appended to the existing context by default.
This implementation balances context reuse with the flexibility to edit the context.

\textbf{Multi-agent extensions of \oursplural.}
Our implementation naturally extends to multi-agent workflows by allowing multiple context files to coexist and remain synchronized with their respective LLM servers.
For example, an \clmhl{agent swarm} can be implemented by initializing the workspace with multiple context files, while \clmhl{subagents} can be initialized and terminated by creating and deleting additional context files.

\providecolor{clmblue}{HTML}{0668E1}
\providecolor{clmactrule}{HTML}{BFC4C9}
\providecolor{clmactgrey}{HTML}{6B7075}
\providecolor{clmactfade}{HTML}{8D939A}
\providecolor{clmactscaf}{HTML}{B9BEC4}

\lstdefinestyle{clmactstyle}{
  basicstyle=\ttfamily\scriptsize,
  columns=fullflexible,
  keepspaces=true,
  showstringspaces=false,
  breaklines=true,
  breakindent=0pt,
  breakautoindent=false,
  postbreak=\mbox{\textcolor{clmactgrey}{$\hookrightarrow$}\space},
  frame=single,
  framerule=0.4pt,
  rulecolor=\color{clmactrule},
  framesep=4pt,
  xleftmargin=0pt,
  xrightmargin=0pt,
  aboveskip=3pt,
  belowskip=1pt,
  escapeinside={(*}{*)},
  moredelim=[is][\color{clmblue}]{@}{@},
  moredelim=[is][\color{clmactfade}]{~}{~},
  moredelim=[is][\color{clmactscaf}]{!}{!},
}

\providecommand{\clmactcap}[2]{%
  \par\noindent\footnotesize
  \parbox[t][2\baselineskip][t]{\linewidth}{\footnotesize
    \textcolor{clmblue}{\textbf{#1}}\ \textbf{#2}}\par%
}
\providecommand{\clmactsrc}[1]{%
  \par\noindent{\scriptsize\textcolor{clmactgrey}{#1}\par}%
}
\providecommand{\clmactcolw}{0.487\linewidth}
\providecommand{\clmactlabel}[2]{\setcounter{subfigure}{#1}\phantomsubcaption\label{fig:clm-examples-#2}}

\begin{figure}[t]
\centering

\begin{minipage}[t]{\clmactcolw}
\clmactlabel{0}{a}
\clmactcap{(a)}{\ours{} builds in-context scoreboards and trackers to orchestrate and monitor subagents with in-place editing.}
\begin{lstlisting}[style=clmactstyle]
!open(p,"w").write("""[[CTX_TURN 1 role=assistant]]!
## STATE (*\textemdash{}*) Erd(*\H{o}*)s Minimum Overlap Problem (compact)
@LEDGER TOP:@ 0.9992491 ...
@AGENTS:@ 21 launched, 5 currently running ...
@KEY FILES:@ /workspace/subctx_4/h_best_final.npy ...
@FINDINGS:@ All methods plateau at 0.381157 ...!""")!
\end{lstlisting}
\clmactsrc{Erd\H{o}s minimum overlap, step 455}
\begin{lstlisting}[style=clmactstyle]
!new="""!@## ORCHESTRATOR STATE (compact)@
@Budget:@ 7/100 used. All 5 slots BUSY (subctx_0..4 ...
@Dead ends:@ simple grid(0.822), hexagonal(0.9977) ...
@Next:@ score my own candidates while workers run ...!"""!
\end{lstlisting}
\clmactsrc{Circle packing $N=26$, steps 194, 198}

\vspace{4pt}

\clmactlabel{2}{c}
\clmactcap{(c)}{\ours{} writes for loops to remove past irrelevant search results or compact overly long outputs when creating a new view.}
\begin{lstlisting}[style=clmactstyle]
@for t in turns[1:]:@ !...!
    !result += f'\n[[CTX_TURN search]]\n!@Searched: {m.group(1).strip()}.@ @No relevant results.@!\n'!
\end{lstlisting}
\clmactsrc{BrowseComp-Plus, step 1509}
\begin{lstlisting}[style=clmactstyle]
@while i < len(lines):@ !...!
    !if !@len(body) > 500@!:! !...!
        !if !'bcp_search' in body!:! !...!
            !result.append(f"!@[Searched: {query}]@!")!
        !elif !'bcp_get_document' in body!:! !...!
            !result.append(f"!@[Retrieved doc {docid}]@!")!
    !else:!
        !result.extend(body_lines)!
\end{lstlisting}
\clmactsrc{BrowseComp-Plus, step 13}
\end{minipage}
\hfill
\begin{minipage}[t]{\clmactcolw}
\clmactlabel{1}{b}
\clmactcap{(b)}{\ours{} creates a new role alongside the original template roles for its own internal notes.}
\begin{lstlisting}[style=clmactstyle]
!re.sub(r"!~\[\[CTX_TURN 4 .*?(?=\[\[CTX_TURN 16)~!",!
  !"""[[CTX_TURN 4 !@role=notes@!]]!
!STATUS: ... James Gallagher (docid=58939) ...""")!
\end{lstlisting}
\clmactsrc{BrowseComp-Plus, step 40}

\vspace{4pt}

\clmactlabel{3}{d}
\clmactcap{(d)}{\ours{} defines and reuses a function to conveniently compact old results with a reference to its maintained note.}
\begin{lstlisting}[style=clmactstyle]
!progress = """![Search progress: VERIFIED ... NEXT: ...]!"""!
!s = re.sub(..., !@progress@!, s)!

@def compact_turns(text):@
    !return re.sub(..., lambda m: !m.group(0).split('\n')[0]
        !+ '\n!@[search results - see progress note]@!', text)!
@s = compact_turns(s)@
\end{lstlisting}
\clmactsrc{BrowseComp-Plus, step 109}

\vspace{4pt}

\clmactlabel{4}{e}
\clmactcap{(e)}{\ours{} reproduces effective behaviors from existing baselines, preserving important facts and future TODOs in the summary.}
\begin{lstlisting}[style=clmactstyle]
!re.sub(r"!~\[\[CTX_TURN 2.*~!",!
  !"![SUMMARY: ... @Kader Asmal Excellence Award@
   @launched 2011 by Mrs A Motshekga@ ...]!")!
\end{lstlisting}
\clmactsrc{BrowseComp-Plus, step 20}
\begin{lstlisting}[style=clmactstyle]
!new="""![EXPLORATION LEDGER - @86 scored attempts@ ...
Best score: 0.9931 (sum_radii=2.6177) from ...
@UNTRIED IDEAS (priority order):@
@1.@ Gradient clipping norm=1.0 with 12k steps
@2.@ Try lam=2200+uniform(0,2800) with 12k steps ...!"""!
\end{lstlisting}
\clmactsrc{Circle packing $N=26$, step 332}
\end{minipage}

\caption{\textbf{Qualitative examples of CLM context-management behaviors.}
\oursplural{} treat context as a file and can arbitrarily edit it using general code interface.
}
\label{fig:clm-examples}
\end{figure}

\textbf{Qualitative examples.}
We show qualitative examples of \oursplural{} managing context as a file in Figure~\ref{fig:clm-examples}, revealing both novel context-management behaviors and effective compaction strategies. For multi-agent orchestration, \ours{} maintains an in-context scoreboard and updates agent status through 163 in-place edits while keeping the context at only 6--8K tokens (\subref{fig:clm-examples-a}). It can create new internal roles such as ``notes'' when rewriting its context (\subref{fig:clm-examples-b}), and use loops to remove irrelevant search results or compact overlong observations (\subref{fig:clm-examples-c}). \ours{} can also define and reuse helper functions: in (\subref{fig:clm-examples-d}), it invokes `compact\_turns' 37 times to maintain a progress note while compacting detailed observations. Finally, it reproduces effective compaction behaviors by compressing 21K tokens into answer-relevant summaries or preserving untried ideas for future explorations (\subref{fig:clm-examples-e}).
We collected these examples from the zero-shot \ours{} evaluation experiments in Section~\ref{sec:agentic_benchmarks}.

\textbf{Efficiency metrics for \oursplural.}
A common serving optimization is prefix-cache reuse, in which cached states are reused for matching prefixes, while all tokens from the first prefix mismatch onward must be re-prefilled, as can occur after an in-the-middle edit. 
To account for this, we measure theoretical inference FLOPs using a metric we call \clmhl{prefix-reuse FLOPs} (see Appendix~\ref{app:prefix_reuse_flops} for details). Formally,
\begin{equation}\label{eq:prefix-reuse-flops}
\mathrm{FLOPs}_{\mathrm{prefix\text{-}reuse}}
=
\underbrace{
\mathrm{FLOPs}_{\mathrm{prefill}}
\bigl(\text{unmatched context suffix}\bigr)
}_{\text{tokens from the first prefix mismatch onward}}
+
\underbrace{
\mathrm{FLOPs}_{\mathrm{decode}}
\bigl(\text{generated tokens}\bigr)
}_{\text{new output tokens}}.
\end{equation}

\subsection{In-Context Learning and Reinforcement Learning for \oursplural}
\label{sec:method_learning}

By treating context management as an LM-native capability,
\oursplural can learn better strategies in context or in weights. 

\textbf{\clmhl{Steering} \oursplural with in-context instruction or skill documents.}
Let $\skl$ denote an in-context instruction or skill document.
\oursplural can be steered by
simply providing $\skl$ as additional in-context guidance to the CLM:

\begin{equation}\label{eq:icl_skill}
c_{t+1}
=
f_\theta^{\mathrm{CLM}}(c_t; \skl).
\end{equation}

\textbf{\clmhl{Evolving} \oursplural with an optimization loop.}
\oursplural can also be optimized through textual evolution. For task instance \(x\), let \(\tau(x;\skl)\) be the trajectory induced by Eq.~\ref{eq:icl_skill}, and \(R(\tau)\) a trajectory-level reward. We optimize
\begin{equation}
\skl^*
=
\arg\max_{\skl}
\mathbb{E}_{x\sim\mathcal{D}}
\left[
R\!\left(\tau(x;\skl)\right)
\right],
\end{equation}
while keeping everything else fixed. 
In our implementation, we use a prompt-evolution loop~\citep{agrawal2026gepa}: In each round, the agent produces rollouts on the training split, and a proposer model uses the resulting traces to generate candidate skills. We evaluate these candidates on the development split and select the skill for the next round. After evolution concludes, we evaluate the final selected skill once on the held-out test split.
The optimizer may be either a stronger external model (\textit{assisted evolution}) or the agent model itself (\textit{self-evolution}), allowing context-management skills to evolve in context.

\textbf{\clmhl{Reinforcement Learning} for \oursplural.}
\oursplural can also learn context-management strategies through reinforcement learning and internalize them in model weights. Since context edits change the input across turns, we use stepwise GRPO~\citep{shao2024deepseekmath}. For each prompt, we sample a group of complete agent trajectories and compute the standard GRPO advantage from their trajectory-level outcome rewards. We then assign each trajectory's advantage to all of its constituent segments, so every model call is trained with the outcome of the full trajectory.

Outcome rewards provide only weak supervision for context editing, as successful trajectories can contain inefficient edits and failed trajectories useful ones.
Simply rewarding edit frequency or removed context volume is also undesirable, as the LM may reward-hack by making unnecessary edits that discard important information or hurt prefix reuse.
We therefore introduce a \clmhl{success-gated efficiency advantage} that further rewards successful trajectories with lower prefix-reuse FLOPs.
Let \(c_i\) denote the prefix-reuse FLOPs of trajectory \(\tau_i\), and let \(\mathcal{G}_g^{+}\) denote the successful trajectories in group \(g\). 
We define
$
\bar{c}_g = \frac{1}{|\mathcal{G}_g^{+}|}\sum_{k\in \mathcal{G}_g^{+}} c_k ,
$
and
\begingroup\small 
\begin{equation}
A_i^{\mathrm{eff}}
=
\begin{cases}
\operatorname{clip}\!\left(
\dfrac{\bar{c}_g - c_i}{\bar{c}_g},
-1,
1
\right), & i \in \mathcal{G}_g^{+}, \\[6pt]
0, & i \notin \mathcal{G}_g^{+}.
\end{cases}
\end{equation}
\endgroup
When there are fewer than two successful trajectories in a group, we set $A_i^{\mathrm{eff}}=0$ for all trajectories. Thus, the efficiency signal only re-ranks among successful trajectories by inference cost.
We combine outcome and efficiency advantages as
$
A_i
=
A_i^{\mathrm{out}}
+
w_{\mathrm{eff}} A_i^{\mathrm{eff}}
$
to encourage trajectories that are both correct and efficient.

\subsection{(More) Efficient \ours Serving with Suffix Cache Reuse}

\begin{figure}[h]
    \centering
    \includegraphics[width=\linewidth]{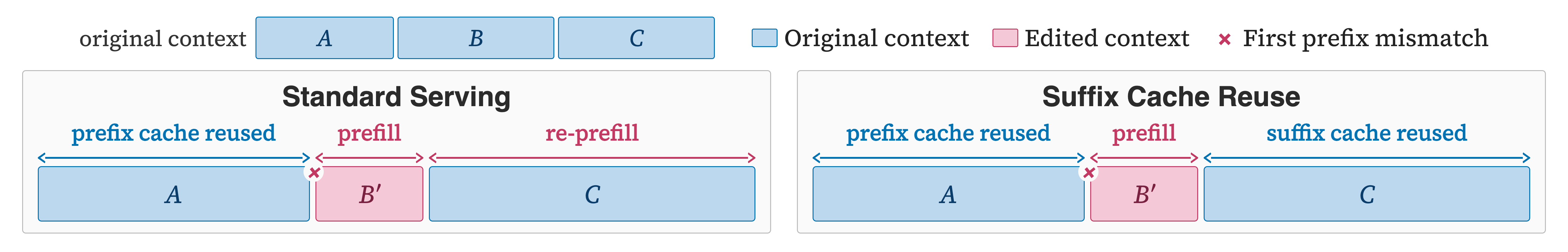}
    \caption{
\textbf{Comparison of standard serving and Suffix Cache Reuse (SCR).} Standard serving reuses only prefix-matched cache, while SCR reuses cached states for all surviving tokens, reducing re-prefilling.}
    \label{fig:scr}
\end{figure}

What if we want to serve \oursplural even more efficiently and reduce the re-prefilling overhead?
We introduce \clmhl{Suffix Cache Reuse (SCR)}. 
As shown in Figure~\ref{fig:scr}, when \ctxorig{B} is replaced by \ctxedit{B'} after an edit, SCR reuses the cached states of all surviving tokens, including \ctxorig{C}, and only reprefills the newly inserted or appended tokens \ctxedit{B'}.
Surviving suffix tokens \ctxorig{C} thus retain stale cache states that encode the previous prefix, which can even be beneficial in some cases, as it retains richer information from the past.
By contrast, standard prefix-cache reuse must re-prefill all tokens after the first mismatch (\ctxedit{B'} and \ctxorig{C}).
\footnote{In some LM chat-serving setups, such as Qwen3.6-27B's default chat template and GPT-5.6 Sol served through the stateless Chat Completions API, reasoning tokens from previous turns are stripped before the next turn, forcing the preserved suffix to be re-prefilled. SCR can also reuse the cached states of these preserved tokens.}
\textbf{Throughout the paper, we report prefix-reuse FLOPs under standard serving; additional SCR savings are reported separately in Section~\ref{sec:scr_results}.}

\section{Results}\label{sec:methods}

\subsection{Evaluating \oursplural{} \clmhl{Zero-Shot} on Long-Horizon Agentic Tasks}
\label{sec:agentic_benchmarks}

We evaluate \oursplural across long-horizon coding, deep research, and open discovery tasks, spanning tens to thousands of agent turns and runtimes from hours to a full day. Our evaluation covers both single- and multi-agent settings, including subagent and agent-swarm workloads for open discovery problems.

\subsubsection{Coding and Deep Research Tasks}
\label{sec:coding_dr}
We first evaluate \oursplural on
two terminal-coding benchmarks, TerminalBench~2.1 (TB2.1)~\citep{merrill2026terminal} and TBLite~\citep{OpenThoughts-TBLite}, and on the deep-research benchmark BrowseComp-Plus (BCP)~\citep{chen2025browsecomp}.\footnote{Deep research requires search tools. Instead of training the agent against a fixed tool interface, we expose the search tools as in-context skills. This follows our less-is-more design principle: we keep as little as possible hard-coded at the harness level, so that tools can be flexibly defined, added, or revised at inference time.}
We compare \oursplural against MEM1~\citep{zhou2026mem1}, Self-Compact~\citep{li2026self}, ACM~\citep{li2026acm}, and recursive language models (RLM)~\citep{zhang2025recursive} with a shared Mini-SWE-Agent backbone~\citep{merrill2026terminal}.
To ensure a controlled comparison independent of training data, we evaluate all methods out of the box without training.
We report performance and prefix-reuse FLOPs on these benchmarks for Qwen3.6-27B with a 32K context budget, with full details in Appendix~\ref{app:eval_config}.

\begin{figure}[t]
\begin{minipage}[t]{0.56\linewidth}
\vspace{0pt}
\centering
\includegraphics[width=\linewidth]{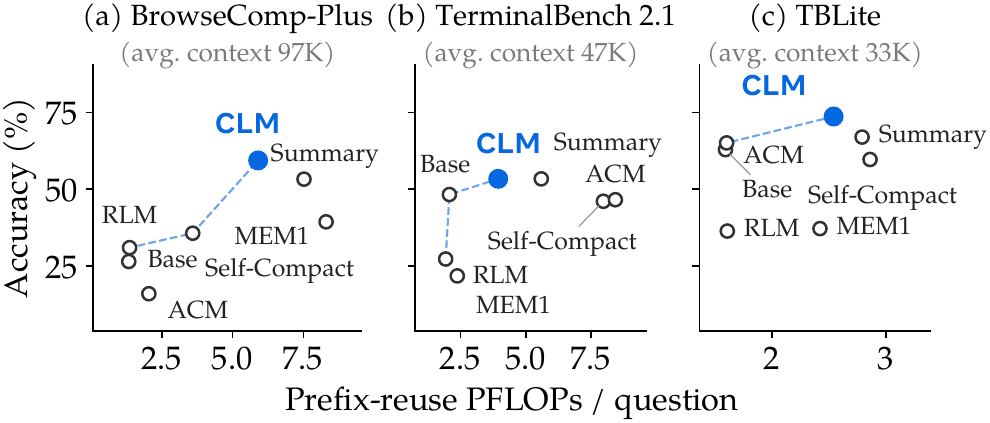}
\caption{\textbf{\oursplural{} perform better than action-based and harness-defined baselines at lower cost on coding and deep research tasks.}
All methods use Qwen3.6-27B with a 32K context limit and a 100-turn cap.
Blue dashed lines indicate the Pareto frontier.
}
\label{fig:pareto_27b}\label{fig:pareto_27b_d}
\end{minipage}\hfill
\begin{minipage}[t]{0.42\linewidth}
\vspace{0pt}
\raggedright
\captionof{table}{\textbf{\oursplural{} outperform specialized OpenEvolve evolutionary workflows on mathematical optimization problems.}
Best-of-run scores with Claude 4.6 Sonnet and a 32K context limit, capped at 100 scored attempts or five hours. Arrows indicate the direction of improvement; OE stands for OpenEvolve and SA for subagents.}
\label{tab:openended-main}
\footnotesize
\renewcommand{\arraystretch}{1.08}
\ifdefined\cptbodylen\else\newlength{\cptbodylen}\fi
\def\cptbody{\toprule
Method & \multicolumn{1}{r}{\shortstack[r]{Circle\\packing ($\uparrow$)}} & \multicolumn{1}{r}{\shortstack[r]{Heilbronn\\($\uparrow$)}} & \multicolumn{1}{r}{\shortstack[r]{Min-max/\\min-dist ($\uparrow$)}} & \multicolumn{1}{r@{}}{\shortstack[r]{Erd\H{o}s\\overlap ($\downarrow$)}} \\
\midrule
OE & 2.541 & 0.03127 & 0.07690 & 0.38123 \\
OE-Agent & 2.525 & 0.03053 & 0.07724 & 0.38167 \\
\rowcolor{swehl} \ours{} & 2.618 & \textbf{0.03653} & \textbf{0.07758} & \textbf{0.38094} \\
\rowcolor{swehl} \ours{} (SA) & \textbf{2.636} & 0.03617 & \textbf{0.07758} & 0.38109 \\
\bottomrule}
\settowidth{\cptbodylen}{\setlength{\tabcolsep}{0pt}\begin{tabular}{@{}l r r r r@{}}\cptbody\end{tabular}}
\setlength{\tabcolsep}{\dimexpr(\linewidth-\cptbodylen)/10\relax}
\begin{tabular}{@{}l r r r r@{}}\cptbody\end{tabular}

\end{minipage}
\end{figure}

\textbf{\oursplural outperform harness-defined and action-based baselines.}
On BCP, \oursplural{} outperform all baselines, scoring 59.4\% at a 32K context limit and exceeding the strongest baseline, Codex-style summarization, by 11.4\% relative. 
\oursplural{} also use 21.5\% and 28.9\% fewer prefix-reuse FLOPs than the next two strongest methods, Codex-style summarization and MEM1, respectively.
On coding benchmarks, \oursplural{} match the strongest baseline, Codex-style summarization, on TB2.1 while using only 70\% of its prefix-reuse FLOPs, and exceed it on TBLite (73.7\% against 67.0\%) with 91\% of its FLOPs.

\subsubsection{Open Discovery Problems}
\label{sec:open_problems_overview}
Open discovery problems provide longer horizons as our testbeds.
We consider three types of open discovery problems with increasing horizons:
(1) \textbf{Mathematical optimization:} four mathematical optimization problems used by AlphaEvolve~\citep{novikov2025alphaevolve} and OpenEvolve~\citep{openevolve}: circle packing, min-max/min-distance 2D, Erd\H{o}s minimum overlap, and the Heilbronn triangle problem.
(2) \textbf{Single-repository optimization:} ten EdgeBench~\citep{zhu2026edgebench} tasks (EdgeBench-10; Appendix~\ref{app:open_single}), where the agent optimizes within a repository for up to 12 hours.
(3) \textbf{Multi-repository optimization with agent swarms:} six repositories jointly optimized by multiple agents and evaluated on held-out downstream packages. Runs last over 24 hours.

\textbf{Mathematical optimization: \oursplural{} vs. specialized evolutionary workflows.}
On mathematical optimization, we compare against OpenEvolve~\citep{openevolve}, a specialized AlphaEvolve-style~\citep{novikov2025alphaevolve} workflow for program generation, evaluation, and evolutionary selection. We also include OpenEvolve-Agent, which replaces its proposer with a Mini-SWE-Agent that can interact with the environment before each submission. 
For \ours{}, we use the same base harness with a minimal Bash interface and provide the evolutionary algorithm as in-context guidance, leaving planning and context management to the agent.
Using Claude 4.6 Sonnet and the same evaluator, \ours{} achieves the highest best-of-run score on all four problems (Table~\ref{tab:openended-main}; progress curves in Figure~\ref{fig:openended-progress}). This shows that a general agent with direct context control can outperform a specialized evolutionary workflow with less fixed orchestration.

\begin{figure}[t]
    \centering
    \captionsetup[subfigure]{justification=centering}
    \begin{subfigure}[t]{0.468\linewidth}\centering
        \includegraphics[height=0.237in]{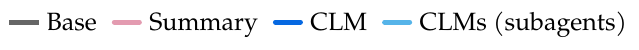}\\[-5pt]
        \includegraphics[height=1.40in]{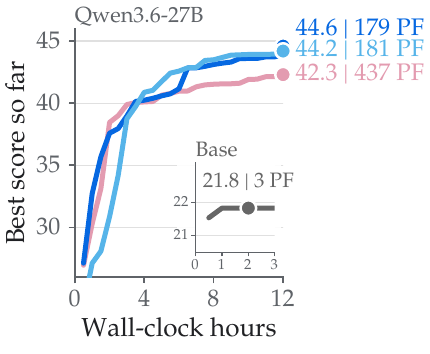}\hspace{-1pt}\includegraphics[height=1.40in]{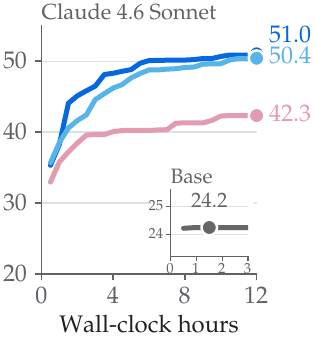}
        \caption{EdgeBench-10 single-repository optimization.}\label{fig:long_horizon_a}\end{subfigure}\hspace{0pt}
    \begin{subfigure}[t]{0.50\linewidth}\centering
        \includegraphics[height=0.237in]{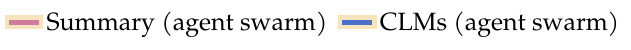}\\[-5pt]
        \includegraphics[height=1.40in]{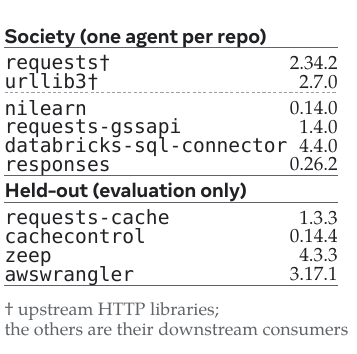}\hspace{5pt}\includegraphics[height=1.40in]{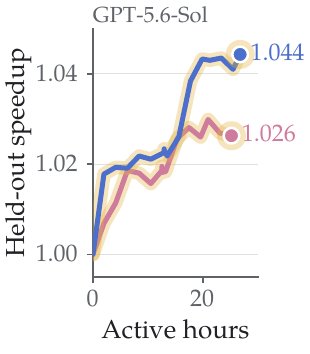}
        \caption{Software World multi-repository optimization.}\label{fig:software_world_b}\end{subfigure}
    \caption{\textbf{\oursplural outperform Codex-style summary harness on long-horizon repository optimization.}
(\subref{fig:long_horizon_a}) 
EdgeBench-10 with a 32K context budget. Curves show best-of-three scores over 12 hours for Qwen3.6-27B and Claude 4.6 Sonnet; end labels show final scores and, for Qwen3.6-27B, mean compute per trial (PF = prefix-reuse PFLOPs).
(\subref{fig:software_world_b}) Software World with GPT-5.6-Sol and a 272K context budget. Six agents jointly optimize interdependent repositories and are evaluated on four unseen downstream packages; the right panel shows geometric-mean speedup over 17 evaluation tasks.}
\vspace{-2mm}
    \label{fig:long_horizon}
\end{figure}

\textbf{Single-repository optimization under single-agent and subagent settings.}
On EdgeBench-10, agents optimize a repository for up to 12 hours with verifier feedback; we report the best score over three seeds per task. Figure~\ref{fig:long_horizon} compares the base harness, Codex-style summarization, \ours{}, and \oursplural{} with up to five concurrent subagents under a 32K context budget. With Qwen3.6-27B, \ours{} reaches 44.6 using 179 prefix-reuse PFLOPs per trial, versus 42.3 and 437 PFLOPs for summarization; the subagent variant reaches 44.2 at 181 PFLOPs. With Claude 4.6 Sonnet, \ours{} and its subagent variant reach 51.0 and 50.4, compared with 42.3 for summarization. 
We find that subagents provide little additional benefit on this single-repository benchmark.

\textbf{Multi-repository optimization with agent swarms.}
We evaluate \ours{} on \texttt{Software World}, where six agents jointly optimize interdependent Python repositories and are evaluated on four unseen downstream packages (Figure~\ref{fig:software_world_b}, left). This provides an extrinsic test of whether improvements transfer beyond the repositories the agents directly observe. Compared with a summary-based agent swarm at the same spend, \ours{} achieves 65\% greater downstream speedup over the initial releases (Figure~\ref{fig:software_world_b}, right). Full setup and scoring details are provided in Appendix~\ref{app:open_multi}.

\subsection{Learning Better Context-Management Strategies in \clmhl{Context} or in \clmhl{Weights}}
\oursplural make context management an intrinsic model behavior that can be learned like other skills. In this section, we present in-context learning and reinforcement learning results for \oursplural.

\begin{figure}[h]
\begin{minipage}[t]{0.60\linewidth}
\vspace{0pt}\centering
\includegraphics[width=\linewidth]{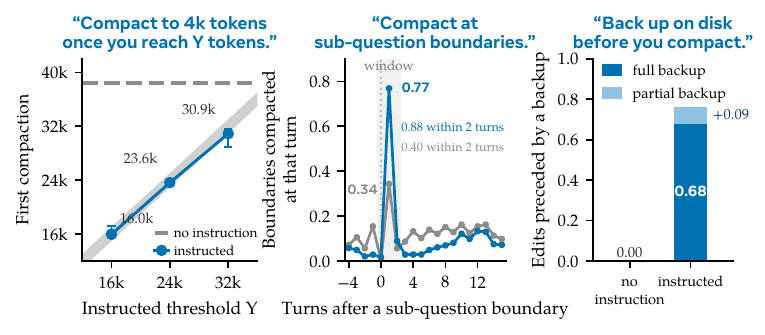}
\captionof{figure}{\textbf{One sentence in the prompt changes the context-management policy.}
Natural-language instructions steer compaction timing, semantic boundaries, and backup behavior. Gray denotes no instruction and blue the instructed setting; exact prompts are in Appendix~\ref{app:steering_setup}.}
\label{fig:steering}
\end{minipage}\hfill
\begin{minipage}[t]{0.38\linewidth}
\vspace{0pt}\centering
\includegraphics[width=\linewidth]{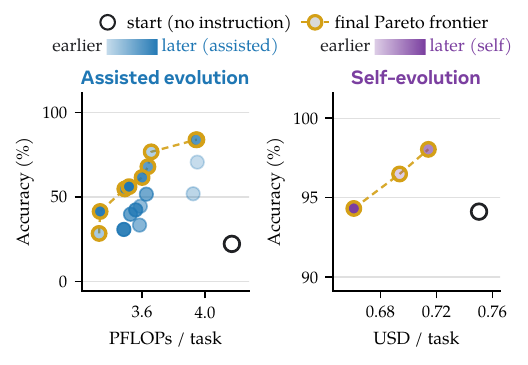}
\captionof{figure}{\textbf{Textual evolution for \oursplural on KV Store (32K budget).}
Assisted evolution uses Qwen3.6-27B with Claude Fable 5.1 as proposer; self-evolution uses Opus 5 for both roles. 
}
\label{fig:selfevo_evolution}
\end{minipage}
\end{figure}

\textbf{Steering context management by simply talking to \oursplural.}
Users can steer \oursplural{} toward a desired context-management strategy through natural-language instructions. We demonstrate this with three behaviors: triggering compaction at a specified context length, compacting around semantic sub-question boundaries, and backing up the context before compaction. Each behavior is induced by a single sentence appended to the task prompt. As shown in Figure~\ref{fig:steering}, the agent adapts its context-management policy accordingly, without any change to the harness or model parameters. Measurement details are provided in Appendix~\ref{app:steering_setup}.

\textbf{Evolving context management via textual evolution with \oursplural.}
We apply the in-context evolution loop from Section~\ref{sec:method_learning} to \ourbench{} with a 32K context budget.
In \textcolor{evoblue}{\textbf{assisted evolution}}, Qwen3.6-27B starts without any context-management instruction, with Claude Fable 5.1 serving as the skill proposer; in \textcolor{evopurple}{\textbf{self-evolution}}, Opus 5 serves as both the agent and the proposer.
Figure~\ref{fig:selfevo_evolution} shows results on KV Store from \ourbench{}, where both settings improve over their initialization and expand the performance--cost Pareto frontier, with evolved skills that can strictly dominate the starting point.
Additional results are in Appendix~\ref{app:selfevo_full}.

\ifdefined\rltabw\else\newlength{\rltabw}\fi
\def\rltabbody{\begin{tabular}{lcc}
\toprule
\textbf{Method} & \textbf{Acc.\ (\%) $\uparrow$} & \textbf{PFLOPs / Q $\downarrow$} \\
\midrule
Summary & 34.7 $\rightarrow$ 42.1 & 4.01 $\rightarrow$ 2.19 \\
\rowcolor{swehl}
\ours{} & 28.8 $\rightarrow$ \textbf{42.5} & 1.52 $\rightarrow$ \textbf{1.34} \\
\bottomrule
\end{tabular}}
\settowidth{\rltabw}{\small\setlength{\tabcolsep}{5pt}\rltabbody}

\noindent\begin{minipage}[t]{\dimexpr\linewidth-\rltabw-1.5em\relax}
\vspace{0pt}
\textbf{Reinforcement learning for \oursplural.}
We post-train Qwen3.5-9B on OpenResearcher using the reward formulation from Section~\ref{sec:method_learning} and evaluate on held-out BrowseComp-Plus. Before training, \ours{} with Qwen3.5-9B underperforms the summary harness by six points due to the smaller model's limited context-management capabilities. After RL, it gains 13.7 points to 42.5\%, matching the trained summary harness while using 1.34 versus 2.19 PFLOPs per question. Adding the efficiency reward further reduces inference cost without a clear loss in accuracy for either \ours{} or the summary harness. Full results and reward ablations are provided in Appendix~\ref{app:rl_results}.
\end{minipage}\hfill
\begin{minipage}[t]{\rltabw}
\vspace{0pt}
\centering
\small
\setlength{\tabcolsep}{5pt}
\rltabbody
\captionof{table}{\textbf{RL results on BrowseComp-Plus with Qwen3.5-9B.}
Performance before and after training on OpenResearcher. The RL checkpoint is selected on a held-out validation set.}
\label{tab:rl_main}
\end{minipage}

\subsection{(More) Efficient Serving with Suffix Cache Reuse}
\label{sec:scr_results}

\begin{figure}[h]
\centering
\includegraphics[width=\textwidth]{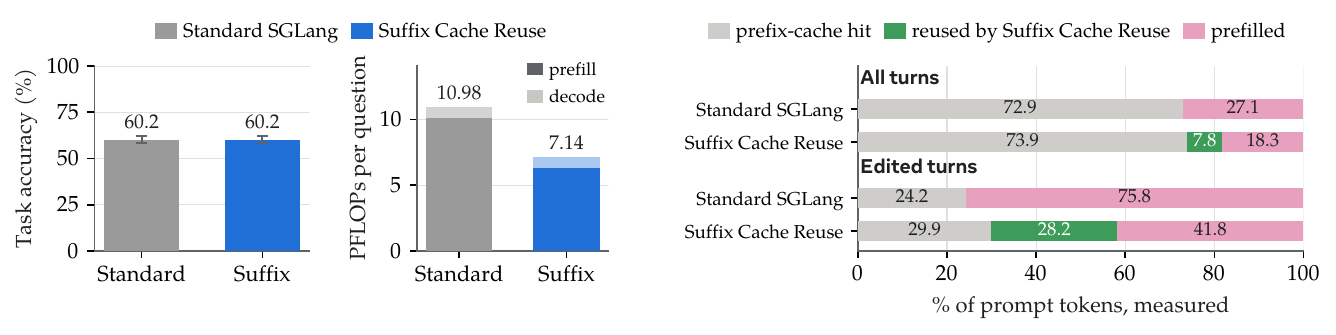}
\caption{\textbf{Comparison of Suffix Cache Reuse and standard SGLang serving on BCP with Qwen3.6-27B.} \textit{Left}: Task accuracy and prefix-reuse FLOPs per question. \textit{Right}: Server-side compute decomposition, showing the fraction of prompt tokens by compute type across all turns and turns following context edits.}
\label{fig:suffix_cache_reuse_wip}
\end{figure}

As shown in Figure~\ref{fig:suffix_cache_reuse_wip}, SCR effectively reduces cache re-prefilling, matching the standard SGLang serving with 65.0\% of its empirical prefix-reuse FLOPs on BCP.
In addition, SCR is not limited to \oursplural. Serving engines commonly strip prior reasoning tokens from chat histories, causing subsequent preserved tokens to be re-prefilled. We show that SCR can also reduce this re-prefilling cost in this more general setting.

We provide further details in Appendix~\ref{app:scr}, including SCR implementation for hybrid models with interleaved full- and linear-attention layers, handling of multiple surviving post-edit spans, a decomposition of savings from reasoning-token stripping, and remaining opportunities for improvement in SGLang serving with SCR.

\section{Discussion and Future Work}

\textbf{Safety implications of a model-editable context.}
Granting models write access to their live context enables more flexible on-the-fly context management, but also creates new safety challenges. Editable context can become another channel through which prompt injections or self-generated instructions persist across turns. Recent work has observed such behavior in compaction summaries, including cases where a model inserted unauthorized instructions into its own summary that subsequently affected task behavior~\citep{openai2026selfgeneratedpromptinjections}. As editable context becomes more widely used, future work should characterize these new attack surfaces and develop defenses that preserve the flexibility of model-controlled context while maintaining its integrity.

\textbf{Future directions: scaling \ours{} RL and distilling existing harnesses into \oursplural.}
Future work can scale RL training so that \oursplural{} can explore and learn effective context-management strategies, and develop a harness-to-\ours{} pipeline that distills strategies from existing harnesses into \oursplural{}. This is motivated by the view that, while standard LMs only map input tokens to next-token distributions, harnesses determine how the context is constructed and updated. Since many harness operations can be expressed as context transformations, they can potentially be translated into \ours{} actions and eventually internalized into model weights. From this perspective, harnesses act as a form of procedural memory or task-specific skill that can be developed externally and later absorbed by \oursplural{} for more general use.

\subsubsection*{Acknowledgments}
We thank Sewon Min and Steven Zijian Chen for helpful discussions. We thank Ilia Kulikov and Mickel Liu for their help with infrastructure questions.
This work was supported by the Singapore National Research Foundation and the National AI Group in the Singapore Ministry of Digital Development and Information under the AI Visiting Professorship Programme (award number AIVP-2024-001) and the AI2050 program at Schmidt Sciences.

\bibliographystyle{assets/plainnat}
\bibliography{paper}

\clearpage
\beginappendix
\section{Extended Related Work}
\label{app:extended_related_work}

\paragraph{Harness-scheduled context management.}
A common approach is to let the harness determine when and how the context is updated. Systems such as Cursor~\citep{cassano2026trainingcomposer}, Codex~\citep{openai2026codex}, and Terminus2~\citep{merrill2026terminal} trigger compaction when the context reaches a predefined length, using a prescribed summarization procedure. MEM1~\citep{zhou2026mem1} instead updates the context at every turn, combining information retained from previous turns with the new observation rather than carrying forward the full history.
Reinforcement learning can improve model performance within these harness-scheduled procedures: Composer~\citep{cassano2026trainingcomposer,chan2026composer}, CompactionRL~\citep{li2026compactionrl}, and MEM1 train models to preserve useful information or continue reasoning effectively under context compaction. Although the resulting context depends on the model's generation, the update schedule and procedure remain prescribed by the harness.

\paragraph{Model control within a constrained action space.}
Another line of work gives the model control over context management through constrained tools that implement predefined strategies, such as compaction, offloading, retrieval, and branching. Self-Compact~\citep{li2026self} and AutoCompact~\citep{zhang2026autocompact} let the model decide when to compact. Context-as-a-Tool~\citep{liu2026context} exposes model-triggered compaction within a structured context workspace, and ACM~\citep{li2026acm} adds offloading and retrieval. AgeMem~\citep{yu2026agentic} combines long-term memory operations with tools for summarizing and filtering the current context. Context Folding~\citep{sun2025scaling} lets the model branch into a sub-trajectory and fold it into a summary upon returning, whereas AgentFold~\citep{ye2025agentfold} condenses recent interactions or consolidates multiple historical steps through folding directives. Sculptor~\citep{li2026sculptor} supports fragment-level summarization, hiding, restoration, and search while preserving message count and order.
Training can improve how models use these tools: AgentFold uses supervised fine-tuning, while AutoCompact, AgeMem, and Sculptor use reinforcement learning to optimize their respective context-management decisions. However, the available operations and their underlying strategies remain predefined by the tool interfaces. 
In contrast, \oursplural{} treat \emph{context as a file}, giving the model direct read and write access to its live context through general-purpose programming tools.

\paragraph{Model-controlled context through meta optimization.}
Meta-optimization gives models control over context management by improving the reusable procedures that govern agent execution. Meta-Harness~\citep{lee2026meta} and AutoMem~\citep{wu2026automem} optimize harnesses or memory-management procedures from trajectory feedback, while Meta Context Engineering~\citep{ye2026meta} co-evolves context-engineering skills and context-construction functions represented as files and code. Related approaches optimize the agent program itself~\citep{zhang2025darwin,zhang2026hyperagents} or evolve prompts through reflection on rollouts~\citep{agrawal2026gepa}.
From the perspective of \oursplural{}, a harness encodes reusable procedures for managing context. Such procedures can also be expressed as skills that guide the model in editing its live context. Our skill evolution therefore shares the goal of harness optimization: improving reusable context-management procedures through evaluation feedback.

\paragraph{Context as a variable in the environment.}
Recursive language models (RLMs)~\citep{zhang2025recursive} place a long input in a REPL variable that the model can access programmatically and process through recursive calls. This gives the model control over how it reads the input, but does not expose its own live context for direct editing. The distinction is twofold: RLMs externalize the input rather than the evolving interaction history, and the model's access to its live context remains read-only rather than read--write. \oursplural{} instead make the live context itself editable, including information accumulated during execution.
The two approaches are complementary: RLM-style access can keep large inputs outside the context until needed, while \oursplural{} can manage the information brought into the context and the history generated while processing it.

\paragraph{Non-prefix KV cache reuse.}
With standard prefix caching, changing an early part of a prompt forces the serving system to recompute the KV states of everything that follows, even when the later text is unchanged. Prior work relaxes this requirement in different settings. Prompt Cache~\citep{gim2024prompt} precomputes attention states for predefined prompt modules, allowing a module to be reused in prompts that do not share the same preceding text. In retrieval-augmented generation, the same document may appear after different documents or instructions. CacheBlend~\citep{yao2025cacheblend} and EPIC~\citep{hu2025epic} reuse cached document chunks in these new contexts, recomputing selected tokens to account for the changed surroundings.
PIE~\citep{he2025let} studies cache reuse when a user modifies previously processed code and requests a new completion. It retains cached states for unchanged text after an edit and corrects their rotary positions, avoiding suffix recomputation. Memento~\citep{kontonis2026memento} evicts each completed reasoning block from the KV cache but keeps the cached states of its summary, which were computed while the block was still in context, and finds that these states retain useful information from the evicted block. Suffix Cache Reuse applies the same reuse principle to an agent's live context: when the agent replaces a span, the unchanged suffix retains its cached states rather than being prefilled again. We integrate this mechanism into SGLang for agent-driven context editing and further extend it to hybrid architectures that combine full-attention layers with linear-attention layers.

\paragraph{Reinforcement learning for context management.}
Context edits break the append-only structure of an agent trajectory: the final context may no longer contain the inputs under which earlier actions were generated. Training must therefore preserve these intermediate contexts and evaluate each generated segment under its original input. ReSum~\citep{wu2025resum} segments trajectories at summarization boundaries and broadcasts the trajectory-level advantage to all segments. Sculptor~\citep{li2026sculptor} similarly preserves training contexts around context modifications and masks previously trained completions to avoid counting them repeatedly. Our stepwise GRPO follows this principle, assigning the final outcome advantage to each segment's policy-gradient loss rather than differentiating through the context edits.
Prior work also considers efficiency. AgeMem~\citep{yu2026agentic} includes a reward for context compactness, while Sculptor penalizes exceeding tool-call or trajectory-length budgets. Our efficiency signal instead measures trajectory-level inference FLOPs under prefix caching, accounting for the recomputation that context edits can incur. We add a success-gated efficiency advantage that favors lower-cost trajectories only within the successful subset of a rollout group. This distinguishes reducing inference compute from merely shortening the context and avoids giving failed trajectories an efficiency bonus.

\paragraph{Success-conditioned efficiency objectives.} Prior work has also conditioned efficiency rewards on task success. \citet{arora2026training} penalize response length only for correct answers, and DDCA~\citep{peng2026think} computes a separate length advantage within the correct-response subset. These objectives primarily measure efficiency through response or trajectory length. Our formulation instead uses trajectory-level FLOPs under prefix caching, capturing the recomputation induced by context edits in addition to generated length.

\paragraph{Context management and external memory.}
Context management determines what the model sees at each invocation, whereas external memory stores information beyond the current context for later use. Memory-R1~\citep{yan2026memory}, for example, learns to manage stored memories and use retrieved information. These mechanisms work together: an agent can offload information to reduce its context and retrieve it when needed. MemGPT~\citep{packer2023memgpt} connects them through a memory hierarchy, allowing the model to edit a designated, fixed-size block within the context rather than the entire live context. AgeMem~\citep{yu2026agentic} jointly learns external-memory operations and tools that summarize or filter the current context.
This distinction depends on the role of the information, not its storage format. In \oursplural{}, the context file specifies the input to subsequent model calls. Other files can serve as external memory, with their contents entering the context when retrieved. Making the live context editable therefore complements external memory by letting the model decide how retrieved information is incorporated and when it is removed.

\section{Suffix Cache Reuse}
\label{app:scr}

\paragraph{Background: radix tree and prefix cache reuse in SGLang.}
SGLang~\citep{zheng2024sglang} keeps the KV cache of served requests in a radix tree over token sequences: each edge holds a token span and the KV entries computed for it, so requests with a common prefix share one path. A new request walks the tree to its longest matching prefix, reuses the KV entries along that path, and prefills only the remaining tokens; unused nodes are evicted in least-recently-used order. Reuse therefore stops at the first mismatched token. This is exact for append-only histories, since each token's key and value depend on all preceding tokens and, through rotary position encodings, on its absolute position. After an in-the-middle edit, however, every token after the edit is re-prefilled, including text that survived unchanged. For hybrid models, whose linear-attention layers keep a fixed-size recurrent state instead of per-token entries, a match can resume only at a node that stores a state checkpoint. Suffix Cache Reuse keeps this tree as is and extends reuse to the surviving tokens beyond the matched prefix, as described next.

\begin{figure}
    \centering
    \includegraphics[width=\linewidth]{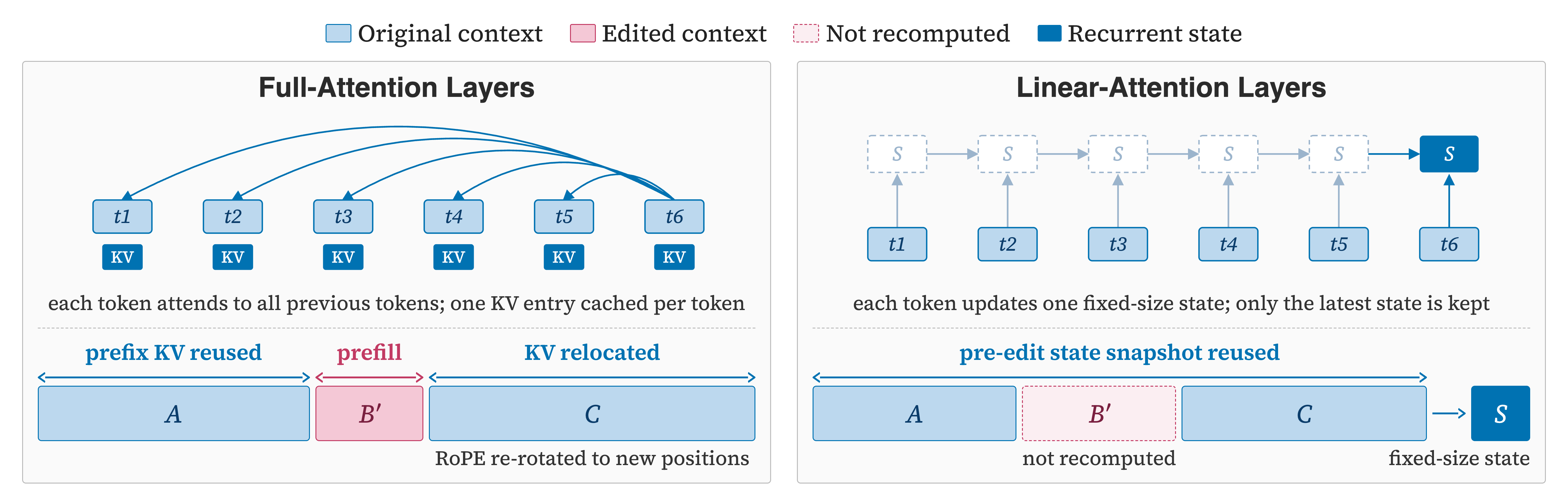}
    \caption{\textbf{Suffix Cache Reuse for full attention layers and linear attention layers.}}
    \label{fig:scr-full-vs-linear}
\end{figure}

\paragraph{Suffix Cache Reuse implementation for standard \clmhl{full attention} layers.}
We implement Suffix Cache Reuse as a patch to SGLang. As shown in Figure~\ref{fig:scr-full-vs-linear} (left), consider a context $[A\,B\,C]$ in which an edit replaces $B$ with $B'$. Standard serving matches only $A$ and re-prefills $B'$ and $C$. When a new prompt arrives, Suffix Cache Reuse diffs it against the session's previous prompt to find the spans that survived the edit, and relocates up to $K$ of them, largest first ($K{=}6$ in the main text). For each relocated span such as $C$, it reuses the cached keys and values, re-rotates the keys' rotary position encodings to their new positions, and splices them in after $B'$. Only $B'$ and newly appended tokens are prefilled. Relocated entries live in session-private cache slots, so the shared radix tree never holds a moved entry; if these slots cannot be allocated, the server falls back to standard re-prefilling. Because the reused states were computed under the pre-edit context, Suffix Cache Reuse approximates re-prefilling, and $K$ bounds the number of relocated spans per edit.

\paragraph{Suffix Cache Reuse implementation for \clmhl{linear-attention} layers.}
In hybrid models such as Qwen3.6-27B\footnote{Qwen3.6-27B is a hybrid model in which 48 of 64 layers use linear attention.}, full-attention and linear-attention layers may be interleaved. As shown in Figure~\ref{fig:scr-full-vs-linear} (right), linear-attention layers maintain a fixed-size recurrent state rather than per-token caches, so there are no token-level entries to relocate. For these layers, we snapshot the recurrent state before the edit and continue from that snapshot, while the edit is reflected only in the 16 full-attention layers that retain token-level context. As a result, the newly inserted \(B'\) is not recomputed in the linear-attention layers, since subsequent tokens depend only on the reused recurrent state. Its representation is still recomputed in the full-attention layers and can influence later linear-attention layers through their inputs.

\paragraph{Suffix Cache Reuse for \clmhl{multiple} surviving post-edit spans.}
A single edit may leave multiple surviving spans after the edit point. SCR can relocate each span, but every relocation reuses states computed under the pre-edit context and therefore introduces an approximation. When many spans are relocated in the same edit, these approximations can compound before the model has a chance to adapt in subsequent turns. We therefore cap the number of relocated spans per edit at \(K\), reusing the \(K\) longest spans and re-prefilling the rest. This bounds the amount of approximation introduced at once and makes SCR less susceptible to pathological edits with many surviving spans.

\begin{figure}[h]
\centering
\includegraphics[width=\linewidth]{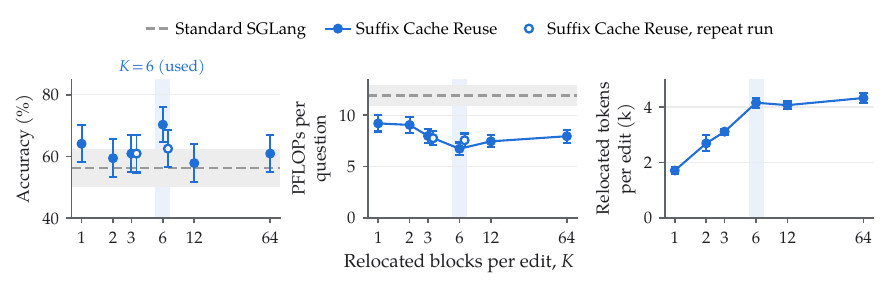}
\caption{\textbf{Small-scale sensitivity study of relocated spans per edit, \(K\).}
Results on 64 BrowseComp-Plus questions with Qwen3.6-27B. Left: task accuracy. Middle: prefix-reuse PFLOPs per question. Right: relocated tokens per edited request. Dashed lines show standard SGLang; hollow markers show a repeat run; the shaded column marks \(K{=}6\), used elsewhere. Error bars show \(\pm1\) standard error across questions; the gray band shows \(\pm1\) standard error for standard SGLang.}
\label{fig:scr_ksweep}
\end{figure}

We conduct a small-scale sensitivity analysis over \(K\in\{1,2,3,6,12,64\}\) on 64 BrowseComp-Plus questions with Qwen3.6-27B (Figure~\ref{fig:scr_ksweep}). Performance is robust across \(K\), while cache-reuse gains largely saturate by \(K=6\). We therefore use \(K=6\) throughout as a conservative choice that captures most reusable cache while limiting the approximation introduced by any single edit. In this small-scale analysis, varying \(K\) mainly affects cache-reuse efficiency, with no observed performance degradation.

\paragraph{Bonus: Suffix Cache Reuse for \clmhl{stripped reasoning tokens} in chat endpoints.}
Chat templates for reasoning models, including Qwen3.6, often remove the reasoning block from earlier assistant turns once the next user message arrives. Standard serving then re-prefills all preserved text after the first removed block, even when the agent never edits its own context. SCR treats reasoning stripping as another context edit and reuses the cached states of the preserved text, extending its benefit to standard chat serving.

\begin{figure}[h]
\centering
\includegraphics[width=\linewidth]{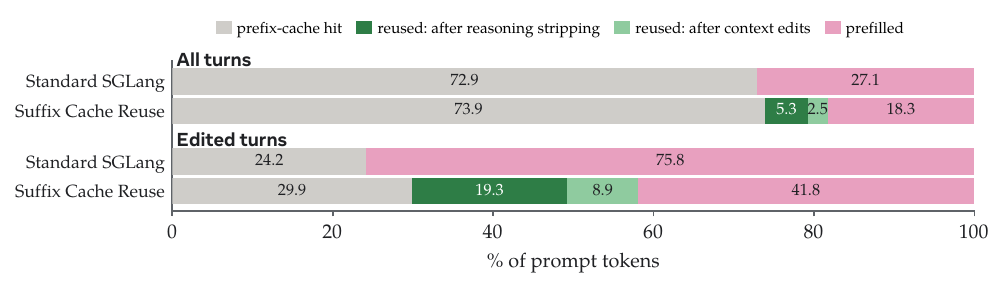}
\caption{\textbf{Suffix Cache Reuse also benefits standard chat serving.}
On BrowseComp-Plus, SCR reuses cached states not only after model-driven context edits but also after reasoning tokens are stripped from prior turns in standard chat endpoints. Results use Qwen3.6-27B on 830 questions with \(K=6\).}
\label{fig:scr_strip_savings}
\end{figure}

Figure~\ref{fig:scr_strip_savings} shows that this effect accounts for a significant portion of SCR's savings on BrowseComp-Plus. Of the 7.8\% of all prompt tokens reused by SCR beyond prefix-cache hits, 5.3 points come from reasoning stripping and only 2.5 from other context edits. Thus, a large fraction of SCR's benefit applies even to standard reasoning-model serving without model-driven context editing.

\begin{figure}[h]
\centering
\includegraphics[width=\linewidth]{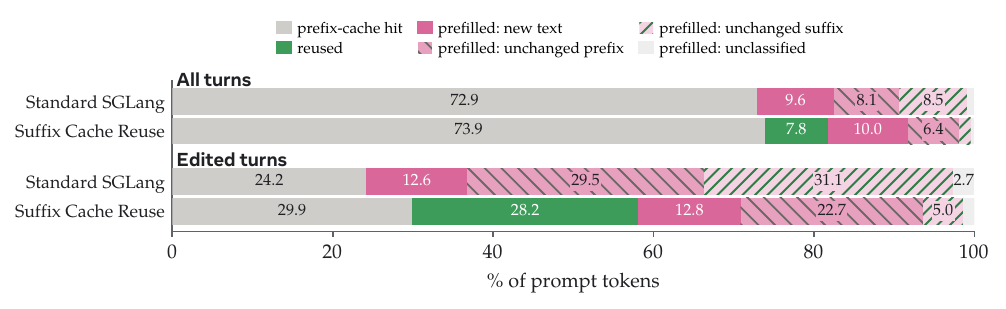}
\caption{\textbf{Remaining re-prefill under Suffix Cache Reuse on BrowseComp-Plus.}
SCR removes most re-prefilling of unchanged suffix tokens, while substantial unchanged-prefix prefill remains due to the current hybrid-model caching behavior in SGLang. Same runs as Figure~\ref{fig:scr_strip_savings}.}
\label{fig:scr_prefill_split}
\end{figure}

\paragraph{A common serving bottleneck and \clmhl{future improvement space}.}
Figure~\ref{fig:scr_prefill_split} shows that SCR removes most re-prefilling of unchanged suffix tokens after an edit. Much of the remaining redundant prefill instead comes from unchanged prefixes that standard prefix caching should ideally reuse. This is a limitation of the current SGLang caching implementation for hybrid models, rather than SCR itself. Linear-attention layers maintain recurrent states, which SGLang stores only at cached request boundaries; when a later prompt diverges inside a cached span, no recurrent state is available near the branch point, so cache matching can fall back to a much shorter prefix. This affects both standard prefix caching and the efficiency attainable with SCR. Storing recurrent states at finer-grained locations, such as message boundaries, is therefore a promising direction for improving both.

\section{Prefix-Reuse FLOPs Computation}
\label{app:prefix_reuse_flops}

Prefix-reuse FLOPs measure the computation performed by a server with prefix caching over an agent trajectory. At turn $t$ the prompt contains $P_t$ tokens and the model generates $G_t$ tokens. The server reuses the longest prefix of leading messages that also appeared in the prompt of an earlier turn, comprising $R_t$ tokens, and prefills only the remaining $U_t=P_t-R_t$ tokens. Under standard prefix caching, an edit therefore invalidates the cached computation from the edited message onward, and a response is prefilled again when it first appears in a later prompt, in addition to being decoded when it is produced.

Consider a model with $L$ layers of hidden size $d$ and MLP width $d_{\mathrm{ff}}$, of which $L_{\mathrm{attn}}$ are full-attention layers with $h_q$ query heads (with an output gate), $h_{kv}$ key--value heads and head dimension $d_h$, and $L_{\mathrm{lin}}$ are Gated DeltaNet layers with $h_k$ key heads, $h_v$ value heads, head dimensions $d_k$ and $d_v$, and two scalar gates per value head. Linear operations have the same cost per processed token regardless of context length. Counting two FLOPs per multiply-add, this per-token cost is
\begin{equation}
C_{\mathrm{token}}
=\underbrace{6Ld\,d_{\mathrm{ff}}}_{\text{MLP}}
+\underbrace{L_{\mathrm{attn}}\bigl[2d(2h_qd_h+2h_{kv}d_h)+2h_qd_hd\bigr]}_{\text{full-attention projections}}
+\underbrace{L_{\mathrm{lin}}\bigl[2d(2h_kd_k+2h_vd_v+2h_v)+2h_vd_vd\bigr]}_{\text{Gated DeltaNet projections}}.
\end{equation}
Only the full-attention layers incur a context-dependent cost. Each query--key pair costs one multiply-add per head dimension for the attention score and one for the weighted value, so the cost per pair is
\begin{equation}
C_{\mathrm{attn}}=4L_{\mathrm{attn}}h_qd_h.
\end{equation}
Ignoring lower-order boundary terms, a turn costs
\begin{equation}
F_t
=C_{\mathrm{token}}\,(U_t+G_t)
+C_{\mathrm{attn}}\Bigl[\tfrac12\bigl(P_t^2-R_t^2\bigr)+G_tP_t+\tfrac12G_t^2\Bigr],
\end{equation}
where the first term inside the brackets counts attention during prefill and the latter two count attention during decoding. The cost of a trajectory of $T$ turns is $\sum_{t=1}^{T}F_t$. We omit the embedding and output layers, the Gated DeltaNet recurrent-state update, and normalization, activation and softmax operations.

\paragraph{Example: Qwen3.6-27B.} Qwen3.6-27B has $L=64$ layers with $d=5120$ and $d_{\mathrm{ff}}=17{,}408$. Its $L_{\mathrm{attn}}=16$ full-attention layers have $h_q=24$, $h_{kv}=4$ and $d_h=256$, and its $L_{\mathrm{lin}}=48$ Gated DeltaNet layers have $h_k=16$, $h_v=48$ and $d_k=d_v=128$. Substituting these values gives $C_{\mathrm{token}}=48.70\times10^{9}$ FLOPs per token, of which the MLP accounts for $34.23\times10^{9}$, the full-attention projections for $3.36\times10^{9}$ and the Gated DeltaNet projections for $11.12\times10^{9}$, and $C_{\mathrm{attn}}=3.93\times10^{5}$ FLOPs per query--key pair.

\begin{figure}[h]
\centering
\includegraphics[width=\linewidth]{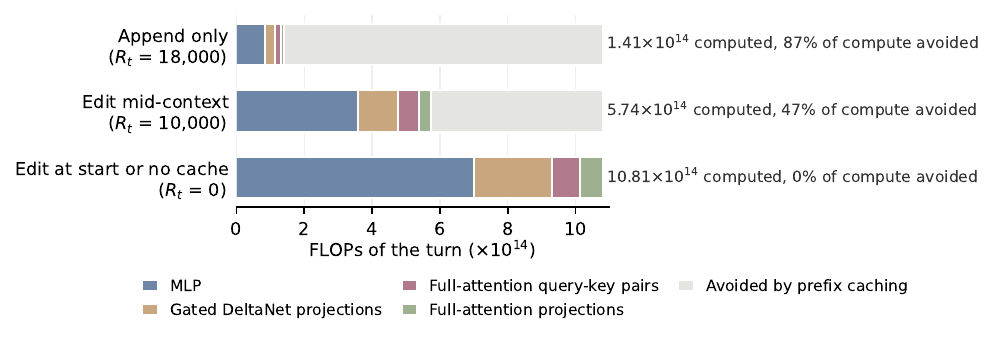}
\caption{FLOPs of one Qwen3.6-27B turn for illustrative lengths ($P_t=20{,}000$, $G_t=500$) and three reusable-prefix lengths $R_t$. Colors split the FLOPs actually computed by operation; gray shows the additional computation that would be required without prefix caching.}
\label{fig:flops_cache}
\end{figure}

Figure~\ref{fig:flops_cache} illustrates the computation saved by prefix caching within one turn. \textbf{The lengths are chosen for illustration and do not come from a particular run}: a prompt of $P_t=20{,}000$ tokens and a response of $G_t=500$ tokens, with the reusable prefix $R_t$ set to 18{,}000, 10{,}000 or 0 tokens. When the turn only appends to its context, prefix caching avoids 87\% of the computation the turn would need without a cache, and the turn costs $1.41\times10^{14}$ FLOPs. An edit in the middle of the context leaves 10{,}000 reusable tokens and raises the cost to $5.74\times10^{14}$ FLOPs. An edit at the start, equivalent here to having no reusable prefix, costs $10.81\times10^{14}$ FLOPs, 7.7 times the append-only turn.

\section{\ourbench}
\label{app:context_bench}

\paragraph{Design and examples of \ourbench tasks.}
Each \ourbench{} task grades what the agent kept, updated, or offloaded from an input that, at all but the lowest levels, exceeds the context window. The environment delivers the input as a sequence of operations, and each operation arrives as a new user message in one continuous conversation. An operation therefore enters the agent's context before the agent can act on it, and no harness can truncate or offload it on the way in. The agent controls only when the next operation arrives: it runs \texttt{echo READY\_FOR\_NEXT\_OP} once it has handled the current one. The tasks need no search or inference beyond following the instruction, so an agent that could keep everything would answer every query; performance measures how the agent manages its context and nothing else. Every task instance is generated from a seed and checked before use. Of the 32{,}768-token context limit, 2{,}048 tokens are reserved for the model's response, leaving a usable budget of 30{,}720 tokens; a single operation must fit in a fifth of it and everything the task requires the agent to retain in half of it, each with a 10\% margin, so a failure at any level reflects how the context was managed rather than a task that cannot be solved within the budget. Table~\ref{tab:ourbench_tasks} summarizes the four tasks and Figure~\ref{fig:ourbench_examples} shows operations from each.

\begin{table}[h]
\centering
\footnotesize
\setlength{\tabcolsep}{4pt}
\begin{tabular}{@{}l>{\raggedright\arraybackslash}p{4.4cm}>{\raggedright\arraybackslash}p{3.3cm}>{\raggedright\arraybackslash}p{4.0cm}@{}}
\toprule
\textbf{Task} & \textbf{Input stream} & \textbf{Queries} & \textbf{Metric} \\
\midrule
Needle Retention & $\sim$4K-token chunks, each with 2--8 needle lines and 140 filler lines & none & needle lines retained verbatim in the final context \\
Sudoku Sketchpad & one move per turn on a $16{\times}16$ board & current board after each move & board versions reproduced exactly \\
KV Store & batches of 100 \texttt{SET} operations with random 24-word values & 24 \texttt{GET} queries & exact-value accuracy \\
Log Triage & batches of 14--54 service log lines & 24 lookup and count queries & exact-answer accuracy \\
\bottomrule
\end{tabular}
\caption{The four \ourbench{} tasks. All metrics are computed from the agent's context; answers held only in files are not credited.}
\label{tab:ourbench_tasks}
\end{table}

\begin{figure}[h]
\centering
\includegraphics[width=\linewidth]{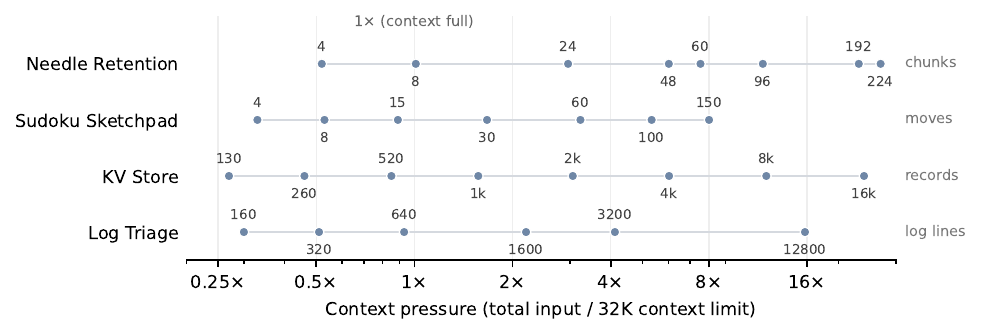}
\caption{Context pressure of every \ourbench{} level. Each row varies one generator setting (unit at right); the label on each point is its value. Total input is every message the environment pushes over the episode, including the task instruction and, for Sudoku, the board once per move, counted in o200k tokens; pressure is total input divided by 32{,}768.}
\label{fig:ourbench_pressure}
\end{figure}

Context pressure is the total input the environment pushes over an episode divided by the 32{,}768-token context limit, so $1\times$ is the point where an agent that keeps everything would fill its context. Each task reaches higher pressure through one generator setting, with everything else fixed (Figure~\ref{fig:ourbench_pressure}); levels below $1\times$ are included as controls on which keeping everything still fits.

\begin{figure}[p]
\centering
\begin{minipage}{\linewidth}
\lstset{basicstyle=\ttfamily\scriptsize,breaklines=true,columns=fullflexible,frame=single,framesep=3pt,xleftmargin=4pt,xrightmargin=4pt,moredelim=[is][\color{clmblue}]{@@}{@@}}
\textbf{\footnotesize Needle Retention} (first of 4 chunks)
\begin{lstlisting}
=== chunk 1/4 (3 op(s) remaining after this) ===
NEEDLES (keep these lines verbatim in your context):
[n00000i00#f56502a0] glacier orbit lantern lattice quartz ember ember meadow lattice orbit.
...
<<<FILLER-BLOCK 00000#d4aff96b START -- delete this entire block (through its END line) from your context this turn>>>
[f00000x000#...] ...   (140 filler lines)
<<<FILLER-BLOCK 00000#d4aff96b END>>>
\end{lstlisting}
\textbf{\footnotesize Sudoku Sketchpad} (initial board, first move, and expected panel)
\begin{lstlisting}
Your STARTING sketchpad (version 0):
<<<SKETCHPAD BEGIN>>>
VERSION: 0
(1,1,.) (1,2,8) (1,3,.) (1,4,G) (1,5,6) (1,6,7) (1,7,C) (1,8,.) ... (1,14,E) (1,15,.) (1,16,4)
(2,1,.) (2,2,.) (2,3,.) (2,4,E) (2,5,.) (2,6,.) (2,7,B) (2,8,.) ... (2,14,.) (2,15,.) (2,16,6)
...   (16 rows)
(16,1,B) (16,2,.) (16,3,.) (16,4,1) (16,5,7) (16,6,4) (16,7,F) ... (16,15,.) (16,16,E)
<<<SKETCHPAD END>>>

=== board 1 move v1 (3 op(s) remaining after this) ===
Move (board #1): Place 9 in cell r1c1 (row 1 from the top, column 1 from the left).
In your sketchpad, set that cell's tuple value to 9 and increment the VERSION line to 1. Leave every other cell unchanged.

@@<<<SKETCHPAD BEGIN>>>@@
@@VERSION: 1@@
@@(1,1,9) (1,2,8) (1,3,.) (1,4,G) (1,5,6) (1,6,7) (1,7,C) (1,8,.) ... (1,14,E) (1,15,.) (1,16,4)@@
@@...   (rows 2-16 unchanged)@@
@@<<<SKETCHPAD END>>>@@
\end{lstlisting}
\textbf{\footnotesize KV Store} (a batch, the last query, and its answer)
\begin{lstlisting}
=== set batch (keys 0-99) (25 op(s) remaining after this) ===
<<<SET-BATCH 0000#9226d9e1 BEGIN -- offload this whole block this turn>>>
SET K00000 = ember crimson saffron falcon marble quartz thistle mistral cobalt cinder ... #121536ed
...   (100 SET lines)
<<<SET-BATCH 0000#9226d9e1 END>>>

=== get 23 (0 op(s) remaining after this) ===
GET K00114
(Report the value you stored for K00114 as an ANSWER block for that key.)

@@<<<ANSWER key=K00114>>>@@
@@garnet copper meadow mistral spruce pewter cinder pewter cobble saffron glacier ... #d17464a7@@
@@<<<ANSWER END>>>@@
\end{lstlisting}
\textbf{\footnotesize Log Triage} (a batch, the last query, and its answer)
\begin{lstlisting}
=== log batch (lines 0-13) (35 op(s) remaining after this) ===
<<<LOG-BATCH 0000#21e46b32 BEGIN -- offload this whole block this turn>>>
2026-06-20T14:00:01Z [INFO] db-proxy req=49e7a0e2 timeout connection cache miss #f49c3100
...   (14 log lines)
<<<LOG-BATCH 0000#21e46b32 END>>>

=== query 23 (0 op(s) remaining after this) ===
QUERY 23: How many [ERROR] log lines are from service "billing"?
(Answer with an ANSWER block for qid=23.)

@@<<<ANSWER qid=23>>>@@
@@4@@
@@<<<ANSWER END>>>@@
\end{lstlisting}
\end{minipage}
\caption{Operations as the agent receives them, from the smallest level of each task, shortened where marked. \textcolor{clmblue}{Blue} text is the output the agent is expected to produce in response; it is graded from the agent's context. Needle Retention asks for no output: its needles are graded in the final context.}
\label{fig:ourbench_examples}
\end{figure}

\paragraph{Detailed instructions for isolated context-management diagnostics.}
In our pilot study, we provide each harness with a detailed task instruction and a method-specific skill that explains how to use its available tools for context management on that task. This helps isolate context-management capability from differences in task understanding or tool use.
For each task, every method receives the same task instruction, followed by a skill tailored to its context-management mechanism. The task instruction specifies the input stream, turn protocol, answer format, and grading procedure, without giving context-management advice. The method skill explains how to manage context using the tools exposed by that harness, including the relevant commands or tool calls, when they take effect, how to verify them, and common failure modes. We write each skill to a level of detail comparable to the \ours{} skill, so that baseline performance reflects what the harness enables rather than what the model must infer about how to use it. Table~\ref{tab:ourbench_skills} shows excerpts of each method's KV Store skill. Skills are inserted at the same position relative to the task instruction for all methods, and all are released with the benchmark.

\begin{table}[!htbp]
\centering
\scriptsize
\renewcommand{\arraystretch}{1.3}
\begin{tabular}{@{}l>{\raggedright\arraybackslash}p{12.4cm}@{}}
\toprule
\textbf{Method} & \textbf{Excerpt of the KV Store skill} \\
\midrule
\ours{} & \textit{Mechanism:} Your context is mirrored to a file you can edit; changing that file changes what you are holding.\newline \textit{Each batch:} \textcolor{clmblue}{On the SAME turn you read a \texttt{SET-BATCH}, move the whole block out of your context and onto disk with this one command} [a \texttt{python3} heredoc that moves the block to \texttt{/tmp/ctx\_offload/} and leaves one placeholder line].\newline \textit{Each GET:} \textcolor{clmblue}{Recover the value with \texttt{grep -h 'SET <key> =' /tmp/ctx\_offload/*} and then print the ANSWER block.}\newline \textit{Check:} No note, or a token count that did not drop, means the batch is still in your context. \\
\addlinespace
ACM & \textit{Mechanism:} \texttt{manage\_context} takes everything since your previous \texttt{manage\_context} call \ldots\ and replaces it with one message \texttt{[summary\_id: N] <summary>}.\newline \textit{Each batch:} \textcolor{clmblue}{Read the gauge after every batch; when it is above 22,000, call \texttt{manage\_context} on your very next turn, before you release another batch.}\newline \textit{Each GET:} \textcolor{clmblue}{Spend one turn on \texttt{query\_memory(N, ``the SET line for key K00084, verbatim'')}} with N the summary whose range covers the key.\newline \textit{Check:} What confirms a compression is the \texttt{[summary\_id: N]} message and a \texttt{[context: \textasciitilde N/M tokens]} readout lower than the turn before. \\
\addlinespace
RLM & \textit{Mechanism:} The REPL namespace persists: every variable and function you define survives from one operation to the next. That persistence is your store.\newline \textit{Each batch:} \textcolor{clmblue}{On the first operation define the store and the handler once; on every later operation only call them} [a dictionary filled by a regular expression over the \texttt{SET} lines].\newline \textit{Each GET:} \textcolor{clmblue}{A GET is answered by placing the ANSWER block in \texttt{answer[\textquotedbl content\textquotedbl]}}.\newline \textit{Check:} \texttt{print(out)} shows \texttt{stored N keys so far} growing by one batch per SET operation. \\
\addlinespace
Context Folding & \textit{Mechanism:} A branch begins from a copy of a context that already holds every batch and cannot delete anything from MAIN, so MAIN is as full after it as before.\newline \textit{Each batch:} \textcolor{clmblue}{On a \texttt{SET-BATCH}, run one command in MAIN and nothing else: \texttt{echo READY\_FOR\_NEXT\_OP}.} \textcolor{clmblue}{Do not open a branch.}\newline \textit{Each GET:} \textcolor{clmblue}{Answer in MAIN \ldots\ by finding the \texttt{SET <key> =} line in the batch in front of you and printing its value.}\newline \textit{Check:} The \texttt{[context: \textasciitilde N/M tokens]} readout should rise by the size of each batch and by almost nothing else. \\
\addlinespace
Self-Compact & \textit{Mechanism:} Compression happens only when C1=Y, C2=Y, C3=Y, N1=N. Then your whole history is replaced by \ldots\ your summary.\newline \textit{Each batch:} \textcolor{clmblue}{On the SAME turn a \texttt{SET-BATCH} arrives, copy the whole block \ldots\ to its own file.} \textcolor{clmblue}{Answer [the probe] so that compression fires as soon as everything you have seen is on disk.}\newline \textit{Each GET:} \textcolor{clmblue}{\texttt{grep -h \textquotedbl\^{}SET K00084 = \textquotedbl\ /tmp/store/*.txt}}, then print the value in an ANSWER block.\newline \textit{Check:} What confirms a store is the count \texttt{grep -c} prints back, equal to 100. \\
\addlinespace
Summary & \textit{Mechanism:} At three quarters of the budget (about 24.6K of 32,768 tokens) \ldots\ everything except the system prompt and this task message is then replaced by one message.\newline \textit{Each batch:} \textcolor{clmblue}{On the SAME turn a \texttt{SET-BATCH} arrives, copy the whole block \ldots\ to its own file.} \textcolor{clmblue}{Write the summary as a pointer, not an inventory.}\newline \textit{Each GET:} \textcolor{clmblue}{\texttt{grep -h \textquotedbl\^{}SET K00084 = \textquotedbl\ /tmp/store/*.txt}}, then print the value in an ANSWER block.\newline \textit{Check:} What confirms the store is the \texttt{grep -c} count coming back as 100. \\
\addlinespace
Base & \textit{Mechanism:} Every operation arrives as a message in this conversation and stays in your context for the rest of the run; there is no command, tool or edit that takes it back out.\newline \textit{Each batch:} \textcolor{clmblue}{On a \texttt{SET-BATCH}, run one command and nothing else.} \textcolor{clmblue}{Do not copy batches to disk.}\newline \textit{Each GET:} \textcolor{clmblue}{Find the \texttt{SET <key> =} line in the batch that is sitting in your context and print its value verbatim.}\newline \textit{Check:} The \texttt{[context: \textasciitilde N/M tokens]} readout should rise by the size of each batch and by almost nothing else. \\
\bottomrule
\end{tabular}
\caption{Excerpts of each method's KV Store skill. Every skill gives the same four kinds of guidance for its own mechanism: how the mechanism works, what to do with each batch, how to answer a query, and how to confirm that a step worked. \textcolor{clmblue}{Blue} marks the concrete instruction for using the method's own tools; \ldots\ marks omitted text and square brackets paraphrase code.}
\label{tab:ourbench_skills}
\end{table}

\section{Experimental Configurations}\label{app:eval_config}

We use the original baseline implementations when available and run all methods with the model and budget specified for each experiment: the summary harness uses the Codex summarization prompts and compacts at 75\% of the budget; Self-Compact uses the original prompts and self-check rubric and asks the model every two turns whether to compress once the context exceeds 37\% of the budget; RLM runs its released harness; ACM runs its released code, in which the model decides when to manage its context. MEM1 is our re-implementation of its inference loop.
\ours{} receives its editing reminder 2{,}048 tokens before the budget\footnote{In Appendix~\ref{app:context_awareness}, we analyze context-length awareness in existing LMs and find that it degrades at long context lengths. Simple environmental hints substantially improve estimation, so we use them as a temporary augmentation. Future models may acquire stronger context awareness directly or estimate usage on demand through context tools.}, and the base harness has no trigger. 
Unless stated otherwise, token budgets are counted with the o200k tokenizer and all methods are evaluated with the same base LM.

\paragraph{ContextBench.}
Every method uses GPT-5.4 through the API with a 32{,}768-token context limit, of which 2{,}048 tokens are reserved for the response, and receives the skill for its harness (Appendix~\ref{app:context_bench}). Each level is run with four seeds, or eight for selected levels to reduce variance. Results are in Figure~\ref{fig:synthetic_task_results_main}.

\paragraph{TerminalBench 2.1.}
We use the 89 tasks of TerminalBench~2.1, each scored as pass or fail by its verifier. Qwen3.6-27B and Qwen3.5-9B are served with vLLM with thinking enabled and up to 4{,}096 generated tokens per call, with a 32{,}000-token budget; a request that would exceed it ends the run. Every method is limited to 64 turns, each shell command to 180 seconds and each task to four times its default time limit. Whenever a run ends, whether the agent finishes, reaches the turn or time limit, or exceeds the budget, the verifier scores the final state of the container. Results are in Figures~\ref{fig:pareto_27b} and~\ref{fig:pareto_9b_27b}.

\paragraph{TBLite.}
We use the OpenThoughts-TBLite tasks. The model is served with vLLM with up to 2{,}048 generated tokens per call and a 32{,}000-token budget. Turn and time limits and scoring are as for TerminalBench~2.1. Accuracy is the mean verifier reward. Results are in Figures~\ref{fig:pareto_27b} and~\ref{fig:pareto_9b_27b}.

\paragraph{BrowseComp-Plus.}
We use all 830 questions over the fixed BrowseComp-Plus corpus. The search tool returns the top 10 snippets of 512 tokens and the document reader is capped at 8{,}192 tokens. Models are served with vLLM with thinking enabled, temperature 0.7, top-$p$ 0.95 and up to 4{,}096 generated tokens per call, with a 23{,}560-token budget and 100 turns; \ours{}'s editing turns do not count toward the turn limit. When a request would exceed the budget, \ours{} rolls back the last turn and retries, up to six times. An answer is correct only if Qwen3.5-27B, judging at temperature 0 with the BrowseComp-Plus grading template, marks it correct and complete; unanswered questions count as wrong. Results are in Figures~\ref{fig:pareto_27b} and~\ref{fig:pareto_9b_27b}.

\paragraph{Math optimization problems.}
We use circle packing (26 circles in the unit square), Erd\H{o}s' minimum overlap, the min--max distance ratio in the plane, and the Heilbronn triangle problem, with one run per method. The agent writes a program that an evaluator runs and scores. Every method uses Claude 4.6 Sonnet with a 32{,}000-token budget and stops after 100 evaluated attempts or five hours. OpenEvolve runs with a population of 60 in four islands; OpenEvolve-Agent uses a Mini-SWE-Agent proposer with 25 turns per candidate; \ours{} with subagents allows up to five subagents of 40 turns each. Results are in Table~\ref{tab:openended-main} and Figure~\ref{fig:openended-progress}.

\paragraph{EdgeBench-10.}\label{app:open_single}
We use 10 of the 48 runnable public EdgeBench tasks (Table~\ref{tab:edgebench10_tasks}) with three seeds each. A run lasts 12 hours in a container without network access; each submission is graded by the EdgeBench judge on a 0--100 scale, and a run's score is the higher of its best graded submission and the grade of the final repository. Qwen3.6-27B is served with vLLM at temperature 0.7 with up to 8{,}192 generated tokens per call, and Claude 4.6 Sonnet runs through the API. The budget is 32{,}000 tokens; when a request would exceed it, the harness rolls back the last turn, up to 50 times, and no turn limit applies. \ours{} with subagents runs up to six concurrent subagents of 40 turns each. The 128K results in Figure~\ref{fig:long_horizon_128k_r50} use a 128{,}000-token budget. Cost is prefix-reuse FLOPs per run, including the subagents' computation. Results are in Figures~\ref{fig:long_horizon}(\subref{fig:long_horizon_a}) and~\ref{fig:long_horizon_128k_r50}.

\begin{table}[h]
\centering
\scriptsize
\setlength{\tabcolsep}{4pt}
\begin{tabular}{@{}lll>{\raggedright\arraybackslash}p{4.2cm}r@{}}
\toprule
\textbf{Task} & \textbf{Category} & \textbf{Language} & \textbf{What the verifier scores} & \textbf{Start: files / LOC} \\
\midrule
\texttt{ad\_placement\_optimization} & Combinatorial optimization & C++ & solution score over judge cases & 1 / 20 \\
\texttt{apple\_incremental\_game} & Combinatorial optimization & Python & solution score over judge cases & 3 / 101 \\
\texttt{graph\_node\_classification} & Science \& ML & Python & held-out accuracy (CPU-only judge) & 2 / 418 \\
\texttt{grid\_turing\_robot} & Combinatorial optimization & Python & solution score (lower is better) & 3 / 440 \\
\texttt{juliet\_vulnerability\_analyzer} & Software engineering & Python & hidden evaluator on the Juliet suite & 1 / 9 \\
\texttt{openrct2\_theme\_park\_ai} & Games \& simulators & JavaScript plugin & park value in a headless OpenRCT2 run & 6 / 5{,}950 \\
\texttt{schemathesis\_datagen\_pipeline} & Software engineering & Python & fraction of the test suite passing & 416 / 112{,}663 \\
\texttt{triangulation\_coloring\_optimization} & Combinatorial optimization & Python & coloring ``ugliness'' (lower is better) & 5 / 431 \\
\texttt{vehicle\_routing\_time\_windows} & Combinatorial optimization & C/C++ & CVRPTW solution quality & 3 / 513 \\
\texttt{wesnoth\_tactical\_ai} & Games \& simulators & Python & headless Wesnoth matches & 0 / 0 \\
\bottomrule
\end{tabular}
\caption{\textbf{EdgeBench-10 tasks.} Scores are rescaled to 0--100 by the EdgeBench judge. Start: files and lines of code in the working directory before any edit.}
\label{tab:edgebench10_tasks}
\end{table}

\paragraph{Software World.}\label{app:open_multi}
Six agents, one per repository (\texttt{requests}, \texttt{urllib3} and four downstream packages), work in parallel for over 24 hours to make their repositories faster. The score is the geometric-mean speedup, in executed instructions, on 17 held-out CPU benchmarks from four downstream packages the agents never see; a benchmark that breaks counts as 1.0. Every agent runs in the Pi agent harness with Pi's default model and context limit, GPT-5.6-Sol with a 272K-token context. Both swarms, with \ours{} and with summary compaction, use long wakes between work sessions. Cost is cumulative API spend in USD. Results are in Figure~\ref{fig:long_horizon}(\subref{fig:software_world_b}).

\paragraph{In-context steering.}\label{app:steering_setup}
All three experiments use BrowseComp-Plus questions with Claude 4.6 Sonnet through the API, the unmodified \ours{} harness, no budget reminders, and a run that ends when a request exceeds the budget. Each behavior is induced by an instruction appended after the task message and compared with a no-instruction control on the same questions; confidence intervals are BCa bootstrap intervals over paired differences. Threshold: a 48{,}000-token budget and 200 turns on 30 long questions (at or above the 60th percentile of turn count without instruction), with the instruction to compact once the context passes 16K, 24K or 32K tokens; we report the median context size at the first compaction. Boundaries: sessions of four chained questions with a 24{,}000-token budget and 1{,}000 turns, 189 sessions per condition, 131 of them with a boundary inside the session; we report the rate of compaction within two turns of a question boundary. Backup: a 16{,}000-token budget on 200 questions, 91 of them paired for analysis; we report the fraction of edits preceded by a full copy of the context. Results are in Figure~\ref{fig:steering}.
The appended instructions are, for the threshold (with $Y\in\{16{,}000, 24{,}000, 32{,}000\}$):
\begin{quotebox}
Monitor your context size: every tool result reports your current size. Whenever it exceeds $Y$ tokens, immediately compact your context down to about 4000 tokens. For example: rewrite \mbox{/tmp/.live\_ctx/LIVE\_CTX\_MAIN.txt} to keep only your key findings and current plan (about 4000 tokens).
\end{quotebox}
for boundaries:
\begin{quotebox}
Please only compress at the boundaries of subtasks. Once a subtask is accomplished, throw out all the context for that subtask. For example: after you write /workspace/answer\_1.txt, rewrite \mbox{/tmp/.live\_ctx/LIVE\_CTX\_MAIN.txt} to drop the Question 1 material, keeping only your answer\_1 conclusion.
\end{quotebox}
and for backup:
\begin{quotebox}
Before every edit to your context, you must back up the current context file into a new file inside a new folder named compaction\_backup; never overwrite or delete earlier backups.
\end{quotebox}

\paragraph{In-context evolution.}
The loop runs on \ourbench{}. Each proposed skill is first run on six training instances; if its accuracy there is at least that of the current frontier, it is run on twelve more, and then on a development split of 102 instances per task with one seed, the only split used to decide frontier membership and selection. For assisted evolution, a test split of 102 instances with three seeds is evaluated once after the archive is frozen. The agent is Qwen3.6-27B served with vLLM, with a 32{,}000-token budget, a 4{,}096-token reserve and 240 turns; a run that exceeds the budget ends. In assisted evolution, Claude Fable 5.1 proposes the skills; in self-evolution, Claude Opus 5 is both the agent and the proposer. Four proposers work in parallel; each reads at least ten rollouts of the current skill, including five successful and failed runs on the same instance, and writes at least four full rewrites, each with a predicted effect on accuracy and cost. A lineage stops after five consecutive proposals without improvement on the development split. Rewards come from the deterministic task graders. A skill enters the Pareto frontier if no earlier skill, the starting point included, is at least as accurate and at least as cheap; the selected skill is the most accurate on the development split, with ties within one standard error broken by cost. Cost is prefix-reuse FLOPs per task over runs that finish within the budget for Qwen3.6-27B, and USD per task for Opus 5. Results are in Figures~\ref{fig:selfevo_evolution} and~\ref{fig:selfevo_evolution_full}.

\paragraph{Reinforcement learning.}
We train Qwen3.5-9B on 3{,}040 OpenResearcher deep-research prompts using GRPO. Training uses truncated importance sampling, a low-variance KL loss with coefficient 0.01, and dynamic sampling that removes groups with no reward variation. Each step samples 8 prompts with 32 rollouts per prompt. We use a learning rate of \(10^{-6}\), weight decay of 0.1, and a clipping range of \([0.2, 0.28]\), and train for 70 steps. Rollouts are served with SGLang at temperature 0.7 and top-\(p\) 0.95, with thinking enabled and up to 4{,}096 generated tokens per turn. \ours{} uses a 28K context budget with a 2{,}048-token reserve and up to 80 turns, excluding editing turns from the count; the summary harness compacts at 28{,}672 tokens and runs for up to 100 turns.
We use a binary task reward from GPT-5.4-nano with the DeepSearchQA rubric. For \ours{}, we additionally apply the efficiency advantage from Section~\ref{sec:method_learning} with weight 0.25 to context-management tokens, together with penalties for failed tool calls and malformed outputs. The summary harness is trained with the task reward alone. Training uses 16 H200 GPUs for the policy and 48 for rollout generation. We select the checkpoint with the highest accuracy on 500 held-out OpenResearcher questions, breaking ties within one standard error in favor of the earlier checkpoint, and evaluate it on all 830 BrowseComp-Plus questions using the same judge. Prefix-reuse FLOPs are computed with Qwen3.5-9B model constants and exact token-level prefix matching against the previous turn. Results are reported in Table~\ref{tab:rl_main} and Figure~\ref{fig:rl_curves}.

\section{Supplementary Results}

\begin{figure}[!htbp]
\centering
\includegraphics[width=0.9\linewidth]{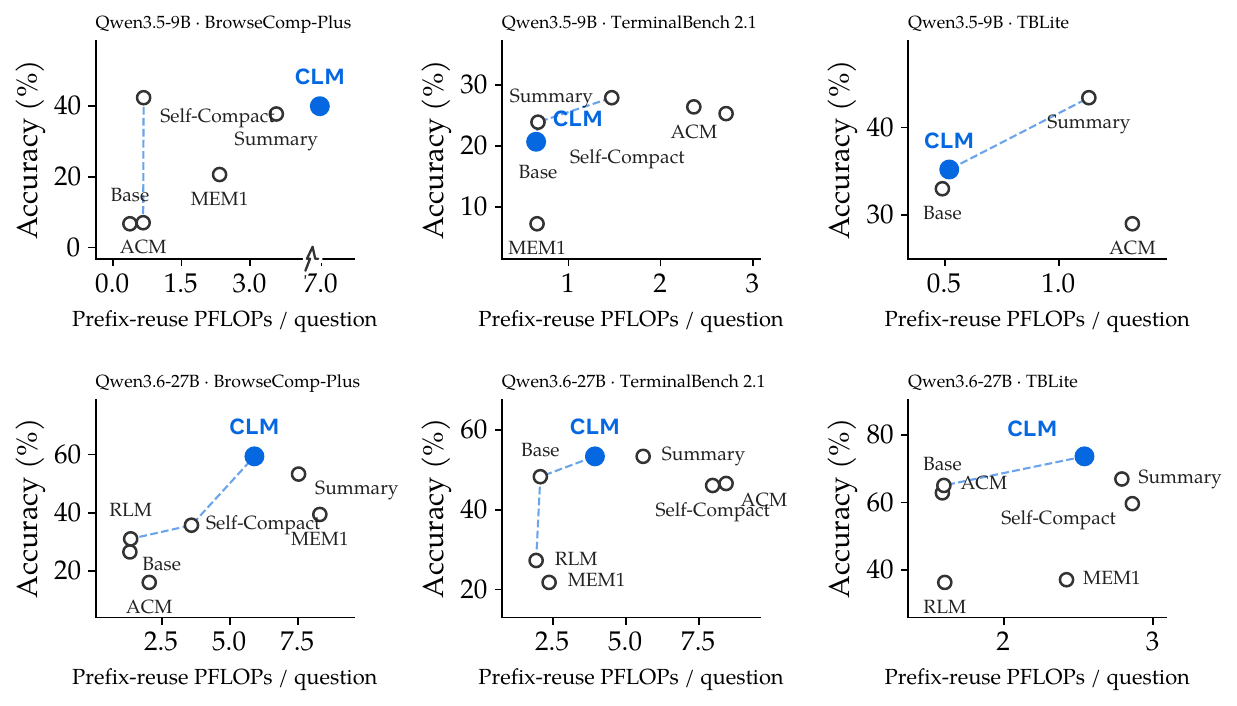}
\caption{\textbf{Qwen3.5-9B and Qwen3.6-27B.} Accuracy against compute for Qwen3.5-9B (top) and Qwen3.6-27B (bottom) on BrowseComp-Plus, TerminalBench~2.1 and TBLite with a 32K context limit. Cost is prefix-reuse PFLOPs per question; dashed lines indicate the Pareto frontier.}
\label{fig:pareto_9b_27b}
\end{figure}

\paragraph{TerminalBench 2.1, TBLite and BrowseComp-Plus with models of different sizes.}\label{app:full_results}
Figure~\ref{fig:pareto_9b_27b} places Qwen3.5-9B next to Qwen3.6-27B on the three benchmarks of Figure~\ref{fig:pareto_27b}. \ours{} leaves the decision of when and how to edit the context to the model, so its gains grow with the model's ability to make that decision. With Qwen3.6-27B, \ours{} lies on the Pareto frontier of all three benchmarks and reaches the highest accuracy on BrowseComp-Plus (59.4\%). With Qwen3.5-9B, \ours{} reaches 39.9\% on BrowseComp-Plus, above the summary harness (37.7\%), and the smaller model edits its context less often: on TerminalBench~2.1, Qwen3.5-9B edits its context 1.4 times per task on average and makes no edit in half of the tasks, while Qwen3.6-27B edits 2.6 times per task. The median peak context is correspondingly higher for Qwen3.5-9B, 30.2K tokens of the 32K limit compared with 17.6K for Qwen3.6-27B.

\begin{figure*}[!htbp]
\centering
\includegraphics[width=\linewidth]{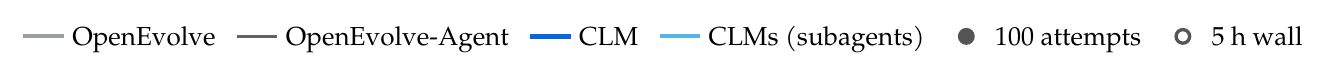}\\[-6pt]
\begin{subfigure}[t]{0.255\linewidth}\centering\includegraphics[width=\linewidth]{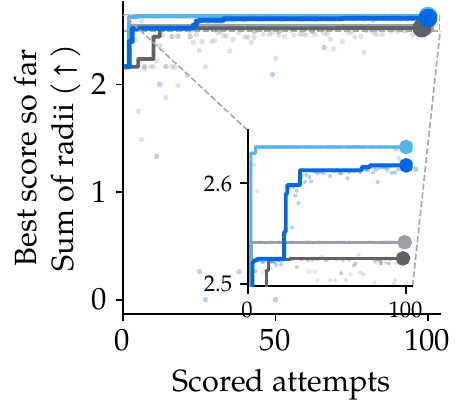}\caption{Circle packing ($\uparrow$).}\label{fig:openended_a}\end{subfigure}\hfill
\begin{subfigure}[t]{0.245\linewidth}\centering\includegraphics[width=\linewidth]{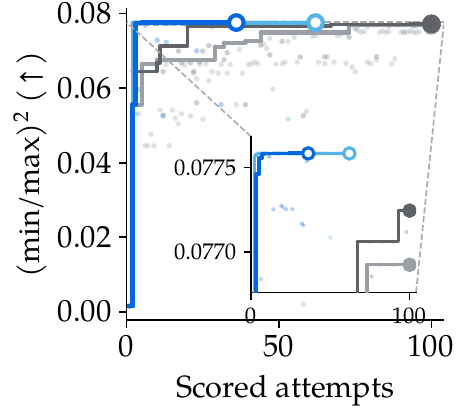}\caption{Min-max/min-dist ($\uparrow$).}\label{fig:openended_b}\end{subfigure}\hfill
\begin{subfigure}[t]{0.245\linewidth}\centering\includegraphics[width=\linewidth]{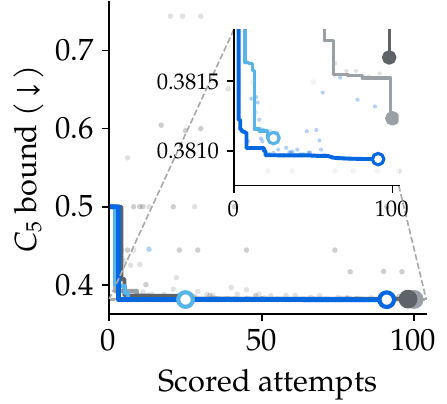}\caption{Erd\H{o}s min-overlap ($\downarrow$).}\label{fig:openended_c}\end{subfigure}\hfill
\begin{subfigure}[t]{0.245\linewidth}\centering\includegraphics[width=\linewidth]{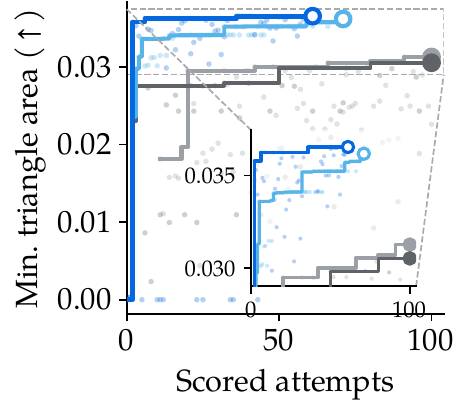}\caption{Heilbronn triangle ($\uparrow$).}\label{fig:openended_d}\end{subfigure}
\caption{\textbf{Best-so-far score versus evaluator-scored attempts on four open optimization problems.}
All runs use Claude 4.6 Sonnet with a 32K context limit and stop after 100 attempts or five hours. Lines show the best score so far, dots individual scored candidates, and insets the final-score range.}
\label{fig:openended-progress}
\end{figure*}

\paragraph{Math optimization problems.}
Figure~\ref{fig:openended-progress} shows the best score so far against the number of scored attempts for each method on the four problems.

\begin{figure}[!htbp]
    \centering
    \includegraphics[width=0.5\linewidth]{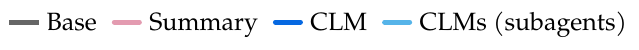}\\[-4pt]
    \begin{subfigure}[t]{0.45\linewidth}\centering\includegraphics[width=\linewidth]{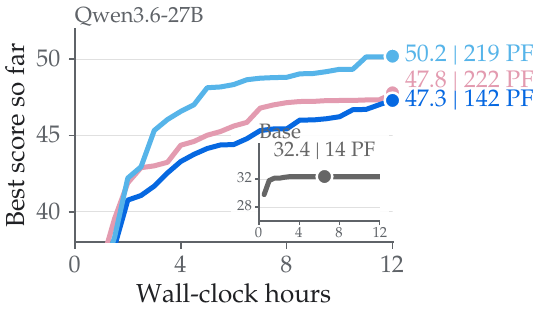}\caption{Score against time.}\label{fig:long_horizon_128k_r50_a}\end{subfigure}\hspace{0.04\linewidth}
    \begin{subfigure}[t]{0.24\linewidth}\centering\includegraphics[width=\linewidth]{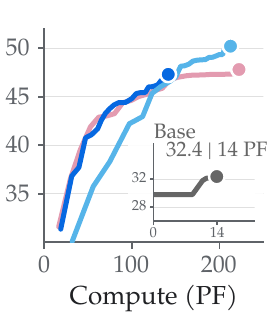}\caption{Score against compute.}\label{fig:long_horizon_128k_r50_c}\end{subfigure}
    \caption{\textbf{EdgeBench-10 single-repository optimization with a 128K context budget.} Qwen3.6-27B, ten tasks, three seeds. End labels give final scores and mean compute per trial (PF = prefix-reuse PFLOPs); insets show the base harness.}
    \label{fig:long_horizon_128k_r50}
\end{figure}

\paragraph{EdgeBench-10 with a 128K context budget.}
Figure~\ref{fig:long_horizon_128k_r50} repeats the EdgeBench-10 comparison of Figure~\ref{fig:long_horizon} with a 128K context budget. The three methods that manage context keep improving over the twelve hours, while the base harness stops improving within the first two hours. At a 32K budget, \ours{} with and without subagents end within 0.4 points of each other (44.2 and 44.6). At 128K, \ours{} with subagents reaches 50.2, compared with 47.3 for \ours{} and 47.8 for summarization, using 219, 142 and 222 prefix-reuse PFLOPs per trial.

\begin{figure*}[!htbp]
\centering
\includegraphics[width=\textwidth]{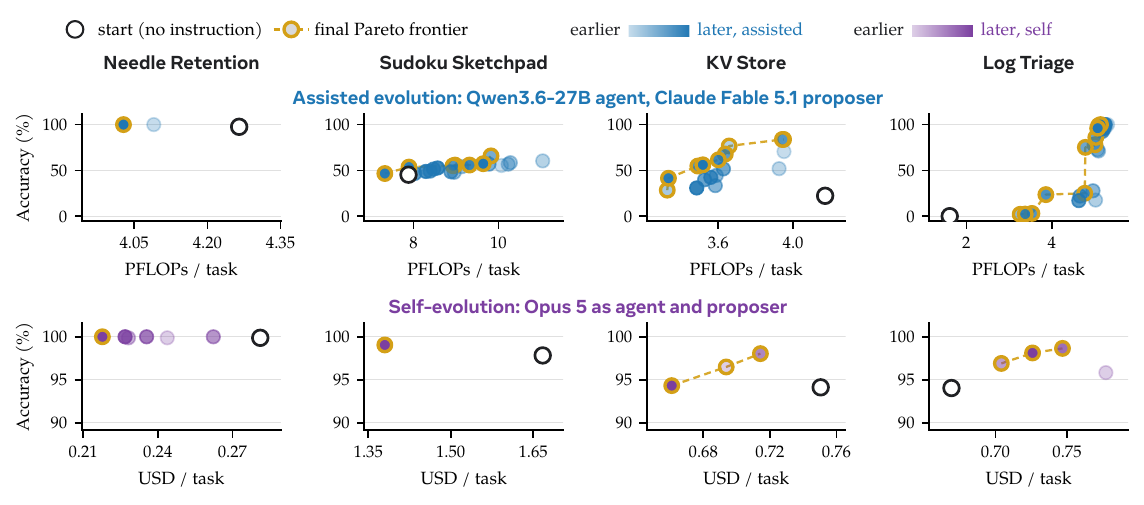}
\caption{\textbf{Evolving context-management skills on \ourbench{} (32K budget), all four tasks.}
\textcolor{evoblue}{\textbf{Assisted evolution}} (top): Qwen3.6-27B serves as the agent, while Claude Fable 5.1 proposes the skills; the x-axis shows prefix-reuse PFLOPs per task over solved runs.
\textcolor{evopurple}{\textbf{Self-evolution}} (bottom): Opus 5 serves as both the agent and the skill proposer; the x-axis shows gateway cost per task in USD.
Skills are proposed based on training-split rollouts and never use the held-out evaluation set.}
\label{fig:selfevo_evolution_full}
\end{figure*}

\paragraph{In-context evolution.}
\label{app:selfevo_full}
Figure~\ref{fig:selfevo_evolution_full} extends Figure~\ref{fig:selfevo_evolution} to all four tasks of \ourbench{}. In assisted evolution, starting without any context-management instruction, the selected skill raises development accuracy from 97.6\% to 100.0\% on Needle Retention, from 45.3\% to 65.8\% on Sudoku Sketchpad, from 22.3\% to 83.8\% on KV Store and from 0.0\% to 100.0\% on Log Triage. On the held-out test split, the selected KV Store skill raises accuracy from 38.3\% to 74.2\%. In self-evolution, Opus 5 starts between 94\% and 100\% accuracy, and the evolved skills either reduce cost at the same or higher accuracy or raise accuracy further.

\begin{figure}[!htbp]
\centering
\includegraphics[width=\linewidth]{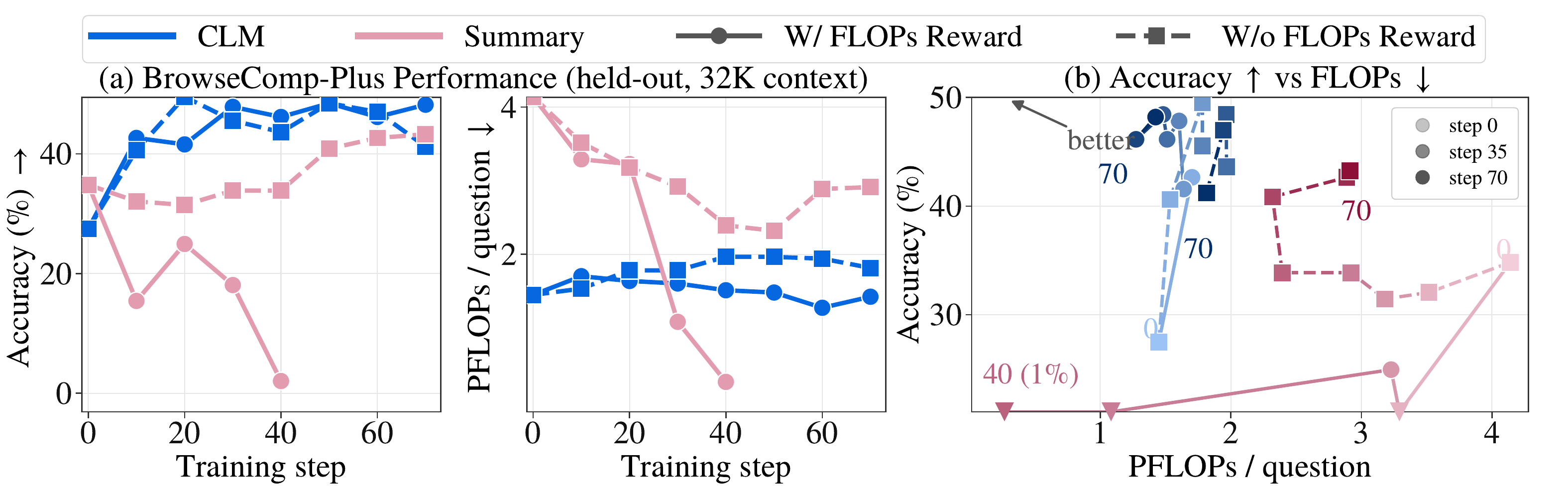}
\caption{\textbf{RL training curves.} Qwen3.5-9B trained on OpenResearcher and evaluated on BrowseComp-Plus with a 32K context limit, through training step 70. (a) Accuracy and prefix-reuse PFLOPs per question against training step. (b) Accuracy against compute, shaded from light to dark by training step.}
\label{fig:rl_curves}
\end{figure}

\paragraph{Reinforcement learning.}\label{app:rl_results}
Figure~\ref{fig:rl_curves} shows accuracy and compute on BrowseComp-Plus during training for \ours{} and the summary harness, each trained with and without the FLOPs reward.

\FloatBarrier
\section{Analysis on the Context Length Awareness of Existing LMs}
\label{app:context_awareness}
An important capability for \oursplural{} is context length awareness: the ability to estimate how much of the context budget has been consumed and to decide when context editing or offloading is needed. Without such awareness, agents may rely on frequent external system interventions, which can leave stale or redundant information in future turns after context editing.

\begin{figure}[h]
    \centering
    \includegraphics[width=\linewidth]{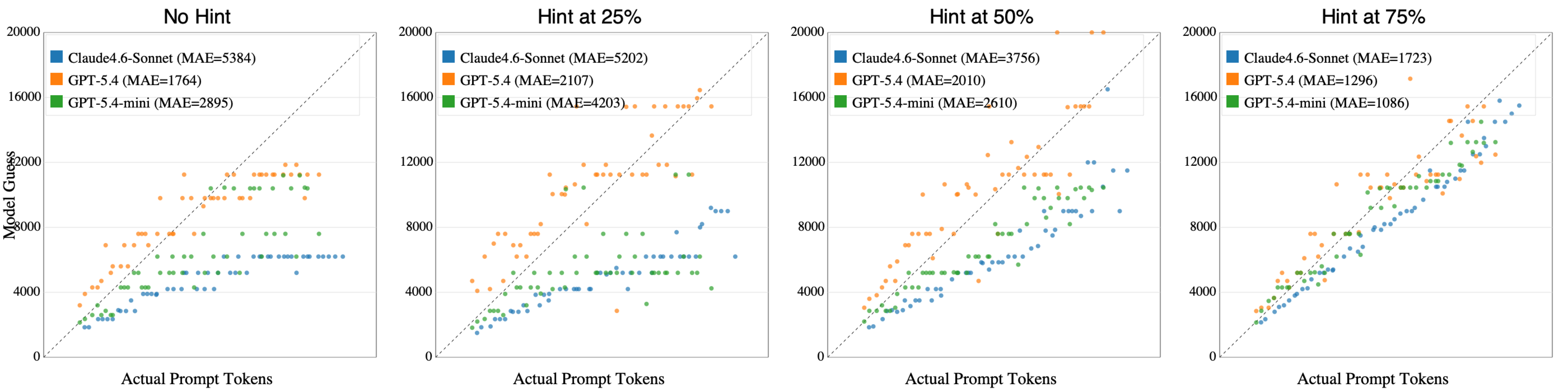}
    \caption{Context length awareness. Each point compares a model's estimated context length with the provider-reported prompt length for the same 50 inputs. Panels vary the available hint: none or a token-count anchor at 25\%, 50\%, or 75\% of the input. The dashed line indicates perfect calibration; legend values report mean absolute error (MAE) in tokens.}
    \label{fig:context_length_awareness}
\end{figure}

We probe context length awareness with a simple diagnostic. We provide each model with prompts of varying lengths and ask it to estimate the number of tokens in the current context. Figure~\ref{fig:context_length_awareness} compares the models' estimated context lengths against the actual prompt lengths. Interestingly, models tend to predict recurring bucketed values in the longer-context regime, such as 6.2K, 9.8K, or 10.4K tokens. We conjecture that these bucketed estimates may reflect token-counting patterns seen during pretraining or post-training. We further observe that Claude-4.6-Sonnet tends to underestimate context length, while GPT-5.4 achieves the strongest alignment between estimated and actual token counts among the three models. Furthermore, token-count hints improve context-length estimation, and hints closer to the estimation point are more effective.
These results indicate that existing LMs have limited context-length awareness at long context lengths, where environmental hints can help substantially. Future \ours{} training may incorporate such signals to improve context-length awareness.

\end{document}